\documentclass{article}

\pdfoutput=1

\usepackage{arxiv}

\usepackage[utf8]{inputenc} % allow utf-8 input
\usepackage[T1]{fontenc}    % use 8-bit T1 fonts
\usepackage{hyperref}       % hyperlinks
\usepackage{booktabs}       % professional-quality tables
\usepackage{amsfonts}       % blackboard math symbols
\usepackage{nicefrac}       % compact symbols for 1/2, etc.
\usepackage{natbib}
\usepackage{graphicx,subfigure}
\usepackage{marvosym}
\usepackage{amsmath}
\usepackage{amssymb}
\usepackage{makecell}
\usepackage{multirow}
\usepackage{enumitem}
\usepackage{tabularx}
\usepackage{CJKutf8}

\title{WakaVT: A Sequential Variational Transformer for Waka Generation}

\author{
	Yuka Takeishi
	\thanks{Yuka Takeishi and Mingxuan Niu contribute equally to this work.}\\
	School of Foreign Studies\\
	Xi'an Jiaotong University\\
	Xi'an, Shaanxi, P.R. China \\
	\texttt{yuka30@stu.xjtu.edu.cn} \\
	\\
	\textbf{Jing Luo} \\
	School of Computer Science and Technology\\
	Xi'an Jiaotong University\\
	Xi'an, Shaanxi, P.R. China \\
	\texttt{luojingl@stu.xjtu.edu.cn} \\
	\And
	Mingxuan Niu
	\footnotemark[1] \\
	School of Computer Science and Technology\\
	Xi'an Jiaotong University\\
	Xi'an, Shaanxi, P.R. China \\
	\texttt{nmx2016@stu.xjtu.edu.cn} \\
	\\
	\textbf{Zhong Jin} \\
	School of Foreign Studies\\
	Xi'an Jiaotong University\\
	Xi'an, Shaanxi, P.R. China \\
	\texttt{jinzhongshici@aliyun.com} \\
	\And
	Xinyu Yang \\
	School of Computer Science and Technology\\
	Xi'an Jiaotong University\\
	Xi'an, Shaanxi, P.R. China \\
	\texttt{yxyphd@mail.xjtu.edu.cn} \\
}

\date{}

\renewcommand{\headeright}{}
\renewcommand{\shorttitle}{}

\hypersetup{
pdftitle={A template for the arxiv style},
pdfsubject={q-bio.NC, q-bio.QM},
pdfauthor={David S.~Hippocampus, Elias D.~Striatum},
pdfkeywords={First keyword, Second keyword, More},
}

\begin{document}
\maketitle

\begin{abstract}
	Poetry generation has long been a challenge for artificial intelligence. In the scope of Japanese poetry generation, many researchers have paid attention to Haiku generation, but few have focused on Waka generation. To further explore the creative potential of natural language generation systems in Japanese poetry creation, we propose a novel Waka generation model, WakaVT, which automatically produces Waka poems given user-specified keywords. Firstly, an additive mask-based approach is presented to satisfy the form constraint. Secondly, the structures of Transformer and variational autoencoder are integrated to enhance the quality of generated content. Specifically, to obtain novelty and diversity, WakaVT employs a sequence of latent variables, which effectively captures word-level variability in Waka data. To improve linguistic quality in terms of fluency, coherence, and meaningfulness, we further propose the fused multilevel self-attention mechanism, which properly models the hierarchical linguistic structure of Waka. To the best of our knowledge, we are the first to investigate Waka generation with models based on Transformer and/or variational autoencoder. Both objective and subjective evaluation results demonstrate that our model outperforms baselines significantly.
\end{abstract}

% keywords can be removed
\keywords{Waka generation \and self-attention mechanism \and variational autoencoder \and linguistic quality \and novelty \and diversity}

\clearpage
\begin{CJK}{UTF8}{min}
\section{Introduction}
\label{sec:1}
Waka\footnote{To avoid ambiguity, our study of Waka is limited to Tanka. For an introduction to various types of Waka, see https://en.wikipedia.org/wiki/Waka\_{}(poetry).} is a type of fixed verse with a long history in Japan. Rich in rhythms and lyricism, Waka can deeply express people’s thoughts and feelings. As one of the most valuable literary genres of classical Japanese literature, Waka is worthy of inheritance and development, which we believe can be further promoted by investigating Waka creation with artificial intelligence (AI). AI has demonstrated its great potential in the area of art creation, such as poetry generation \citep{oliveira2017survey} and music composition \citep{ji2020comprehensive}. However, it still remains a controversial issue whether AI is able to create art like human beings, and presently, poems created by machines cannot really stand comparison with those of human poets. Poem creation is a long-term challenge for AI, while our study on Waka generation can provide reference and enlightenment for the development of this field. In addition, through building intelligent Waka generation systems, we can promote potential applications of humanizing AI in various areas such as electronic entertainment and cultural education.

Due to poetry's attractive aesthetic value, the study of automatic poetry generation has been popular for many years. Among traditional methods, the representative ones are approaches using templates and/or rules \citep{Colton2012FullFace,oliveira2012poetryme}, genetic algorithms \citep{manurung2012using}, statistical machine translation methods \citep{he2012generating}, and text summarization methods \citep{yan2013ipoet}. Traditional methods typically require experts to develop well-designed rules to remedy the defect that the models have no deep understanding of the semantic meaning. With the fast development of deep learning, neural network-based poetry generation systems have been proposed \citep{van2020automatic,zhipeng2019jiuge}. These systems deal with a large amount of linguistic features and semantic relations in poetry more automatically through well-designed network structures and sufficient training data. In the scope of Japanese poetry generation, various Haiku generation systems have been established based on traditional methods \citep{rzepka2015haiku,tosa2008hitch}. In recent years, RNN language models and Generative Adversarial Networks (GAN) have been introduced into the study of Haiku generation \citep{hirota2018haiku,wu2017haiku}.

However, Waka generation is more challenging and less studied. Based on the interactive genetic algorithm, a Waka generation system which is able to process Kansei information given texts from the user has been established \citep{yang2016text}. Masada et al. \citep{masada2018lda} has proposed a scoring method based on the latent Dirichlet allocation (LDA) to select diverse Waka poems generated by the RNN model. Nevertheless, there is still no study that specifically proposes a deep generative model to generate Waka poems with high content quality, as far as we know. The reasons for this include: (1) Compared with Haiku, Waka is written in ancient Japanese and more difficult to comprehend. (2) Waka is short in length but rich in content, and can be used to express various themes (seasons, love affair, life philosophy, etc.), which raises challenges for data modeling. (3) Voiced sounds\footnote{In the Japanese writing system, the \emph{dakuten} (voicing mark) is used to indicate kana characters supposed to be pronounced voiced. However, it's not used by ancient poets in the writing of Waka, making it difficult to recognize the pronunciations. For more information, see https://en.m.wikipedia.org/wiki/Dakuten\_{}and\_{}handakuten.} are hard to recognize in the writing system of Waka, making it difficult to build a large and available Waka dataset of segmented words.

\begin{figure}
	\centering
	\includegraphics[width=5.6in]{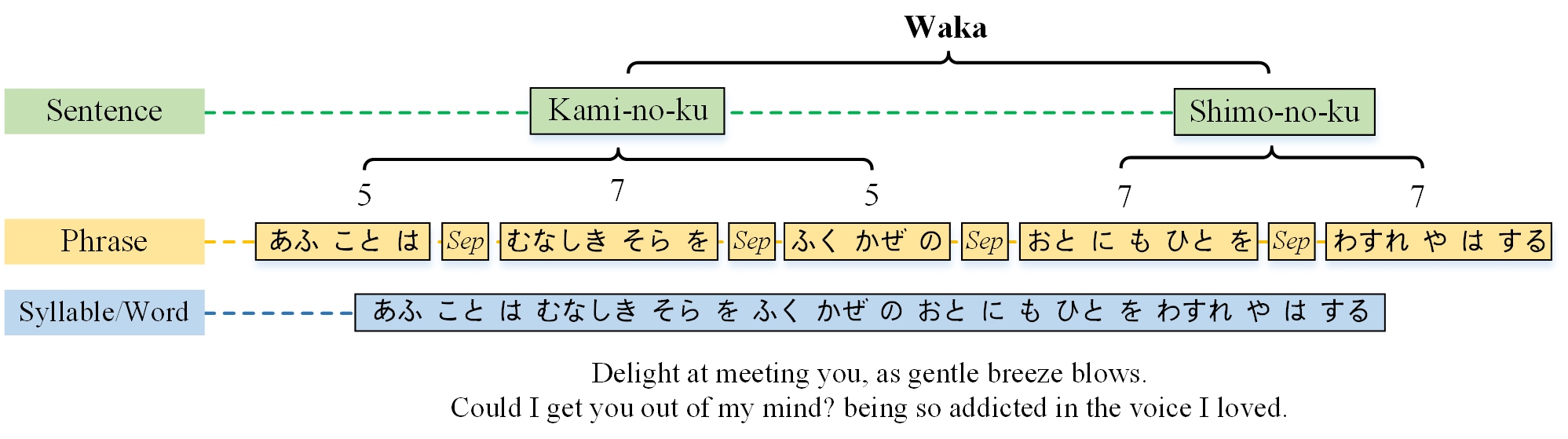}
	% figure caption is below the figure
	\caption{Hierarchical linguistic structure of Waka}
	\label{fig:1}       % Give a unique label
\end{figure}

As is done in general poetry generation tasks, both form and content features need to be considered. In terms of form, Waka follows strict morae constraint, with a short form but a hierarchical structure (Fig.~\ref{fig:1}). Waka can be divided into five phrases following the morae pattern of 5, 7, 5, 7, 7. Generally, the first three phrases and the last two phrases are referred to as Kami-no-ku (or ``upper verse") and Shimo-no-ku (or ``lower verse"), respectively. Morae constraint and hierarchical structure bring formal beauty to Waka, but also present challenges in Waka generation. In terms of content, the linguistic quality, novelty, and diversity of the generated poems are important factors in evaluating the capability of a poetry generation model. For linguistic quality, a generated poem is expected to follow grammatical rules, achieve semantic coherence and convey a meaningful message. As for novelty, the model should create poems itself, rather than copying or recombining pieces from human-created poems. To reflect diversity, poems generated with different inputs should be distinguishable from each other, and repetitive, generic results should be avoided. It is noteworthy that the relationship among the above factors is not simple. Novelty and diversity tend to be positively correlated since both of them reflect the imagination and inventiveness in the selection of content and the use of language. However, they are usually incompatible with linguistic quality as intriguing wording and phrasing may cause problems of linguistics. Consequently, these factors actually need to be balanced in order to get convincing results.

Waka, short in form, requires a highly condensed language to express rich content, thus novelty and diversity become prominent challenging factors. For the generation of other types of poetry, several approaches have been put forward to improve novelty or diversity. Zhang et al. \citep{zhang2017flexible} combined RNN and memory networks to generate innovative Chinese poems. Li et al. \citep{li2018generating} applied CVAE to Tang poetry and Song lyrics generation to improve term novelty. Hirota et al. \citep{hirota2018haiku} proposed using different datasets to pre-train the generator and the discriminator of SeqGAN to avoid plagiarism in the generation of Haiku. Shen et al. \citep{shen2020compose} proposed a novel method of polishing drafts, which improves the novelty and diversity of the language usage in modern Chinese poetry through impressive word detection mechanisms. In the above approaches, CVAE models the text by introducing a single latent variable, which is, however, insufficient to capture the high variability of texts \citep{du2018variational}. In the scope of dialogue generation and machine translation, models based on a sequence of latent variables have been proposed and proven to generate more diverse, informative texts \citep{du2018variational,lin2020variational,schulz2018stochastic}. These models introduce a latent variable for each token of the input sequence, thus capable of modeling the variability at the word level. We will show that it’s feasible to incorporate a sequence of latent variables into Waka generation models to promote their creativity.

In this paper, we propose a Waka generation model, WakaVT, which handles both the form constraint and the content quality. WakaVT is developed by incorporating a sequence of latent variables sequentially into a conditional Transformer language model \citep{keskar2019ctrl,vaswani2017attention}. It takes a user-specified keyword as input through control codes \citep{keskar2019ctrl}. As for form, we propose an additive mask-based method to satisfy the morae constraint of Waka. In terms of content, WakaVT employs a sequence of latent variables to capture the high variability of Waka data to enhance the novelty and diversity of generated content. Besides, we propose the Fused Multilevel Self Attention Mechanism (FMSA) to properly model the hierarchical structure of Waka, enabling a better understanding of the linguistic features, and as a result, improving the linguistic quality of the generated poems. With no prior study for comparison, we build three baselines based on the RNN or the Transformer, considering both language model and variational autoencoder (VAE) architectures. To evaluate the generated Waka in a more comprehensive way, we put forward word-based and 5/7-morae phrase-based automated metrics for novelty and diversity. Both objective and subjective evaluation results demonstrate that WakaVT is able to create Waka poems with strong novelty and diversity, and significantly boosts the linguistic quality in terms of fluency, coherence, and meaningfulness, compared to the baselines.

In summary, our contributions are as follows:
\begin{itemize}
	\item [(1)]
	To the best of our knowledge, this is the first report applying models based on Transformer and/or VAE to the study of Waka generation. The models we built can automatically generate Waka based on user-specified keywords.
	\item [(2)]
	We propose WakaVT, a novel model for Waka generation, which improves novelty and diversity of the generated results through a sequence of latent variables. Moreover, we propose the Fused Multilevel Self Attention Mechanism (FMSA) to boost the linguistic quality by making the model aware of the hierarchical structure of Waka.
	\item [(3)]
	We design several baselines for generating Waka with given keywords. The objective and subjective evaluations indicate that our model outperforms baseline models significantly.
\end{itemize}

The rest of this paper is structured as follows: Sect.~\ref{sec:2} presents previous study related to the work in this paper. Sect.~\ref{sec:3} elaborates on the WakaVT model. Sect.~\ref{sec:4} illustrates the experimental results and analysis. Sect.~\ref{sec:5} draws conclusions.

\section{Related work}
\label{sec:2}
From the perspective of task, our study focuses on the controllable poetry generation given keywords. This section first introduces related studies on controllable poetry generation based on deep learning (Sect.~\ref{sec:2.1}), and then discusses research progress in Japanese poetry generation (Sect.~\ref{sec:2.2}).

\subsection{Controllable poetry generation}
\label{sec:2.1}
Just as a human being is motivated by a certain stimulus to create poetry, a poetry generation system also needs specific prompts to purposefully generate poems. Poetry generation systems are generally conditioned on a user-provided query, so that the generation reflect, to some extent, the user's writing intent. Different types of queries can control different aspects of poetry.

The most common query is to input words or texts to control the poetry's content features. The text provided by the user is often used to guide the system to generate a poem of specific topics, scenarios and concepts. According to the granularity of the input control over the generation, such studies can be divided into three categories. Firstly, some researchers used the keywords to explicitly control the content of the first line of a poem \citep{deng2020iterative,zhang2014chinese}. Zhang et al. \citep{zhang2014chinese} expanded, combined, and selected several keywords entered by the user to obtain the first line of a Chinese poem, and then used RNN to complete it. Deng et al. \citep{deng2020iterative} used the BERT-based Seq2Seq model to convert the given keywords into the first line of a Chinese poem and then completed and polished the draft. Secondly, some researchers used keywords as global information to ensure that the poetry meets the theme expected by the user \citep{ghazvininejad2016generating,li2018generating}. The Hafez system \citep{ghazvininejad2016generating} adopted word2vec to calculate topically related words and phrases, and combined the Finite-state Acceptor and RNN to generate English poetry. Li et al. \citep{li2018generating} took the user-specified title as the topic, and applied a CNN-based discriminator to learn the thematic consistency between the title and each line of the poetry with adversarial training. Thirdly, to have more fine-grained control over the content, researchers used the TextRank algorithm to extract a keyword for each line of the poem from the words, sentences, or documents provided by the user \citep{wang2016chinese,yang2018generating}. Wang et al. \citep{wang2016chinese} proposed a planning-based method inspired by human creation outlines, in which keywords are extracted from the input text, and a sub-topic is assigned to each line of the poem in the planning phase. Yang et al. \citep{yang2018generating} proposed a CVAE model based on a hybrid decoder that effectively enhances the sub-topic information in latent variables utilizing a deconvolutional neural network.

In addition to the above studies, researchers have also explored the possibilities of taking styles, emotions, rhetoric methods, formats, or images as queries for poetry generation. In terms of style, some poetry generation systems could generate poems of specific styles represented by labels \citep{yang2018stylistic,yi2020mixpoet}. In terms of emotion, users can control the emotions expressed in poetry in different ways. For example, users write a blog containing emotional content, which is then machine-converted into English poetry to express their emotions \citep{misztal2014poetry}. Users can also directly input sentiment labels into the model to control the polarity and intensity of sentiment when generating Chinese poetry \citep{chen2019senti}. In terms of rhetoric, Liu et al. \citep{liu2019rhetorically} proposed an encoder-decoder framework which is able to control the use of metaphor and personification in modern Chinese poetry. In terms of format, several unified generation frameworks \citep{zhipeng2019jiuge,hu2020generating} support the generation of poetry of multiple genres and formats. These systems either provide users with multiple options, or define a unified input format to control different genres and formats. The SongNet model \citep{li2020rigid} provides a method for fine-grained control of text formatting, where the user inputs a sequence of placeholders to specify the length, syntactical structure, rhyming scheme, or partial predefined content of the Song Lyrics or Sonnet. In addition, there are also studies on poetry generation that take image data as multi-modal inputs \citep{liu2018multi,xu2018images}, attempting to create poetry related to the object, emotion, scene, or topic of an image.

Our task is to take a keyword provided by the user as a query to make the model generate a Waka poem containing this keyword. Due to the short length of Waka, users are allowed to input only one keyword, whose position in the generated poem is not limited, thus encouraging the model to create concise and novel content with the keyword as the core.

\subsection{Japanese poetry generation}
\label{sec:2.2}
The study of Japanese poetry generation mainly focuses on Haiku, a fixed verse developed from Waka. Haiku is shorter than Waka and follows the morae pattern of 5, 7, 5. The research methods for Haiku generation are generally in three categories: approaches using templates and/or rules, genetic algorithms, and approaches using neural networks. The study on Waka generation is also discussed according to these categories.

The first approach ensure the generated Haiku is formally and grammatically correct through templates or rules, and fills the Haiku with words or phrases extracted from rich lexical resources. The template used can be either an artificially prescribed Haiku format \citep{tosa2008hitch} or a syntactic pattern drawn from the Haiku corpus \citep{netzer2009gaiku,rzepka2015haiku}. The repository of lexical resources is built using thesauri \citep{tosa2008hitch} or web blogs related to Haiku \citep{rzepka2015haiku,wong2008automatic}. In addition, researchers considered various methods to ensure the generated Haiku made sense. Wong et al. \citep{wong2008automatic} calculated the semantic similarity between phrases based on the vector space model (VSM) to match the three lines of a Haiku poem with each other. Netzer et al. \citep{netzer2009gaiku} used the Word Association Norm Network to ensure correlation between the theme words contained in the candidate lines of Haiku. Ito et al. \citep{ito2018haiku} selected and arranged elements from the narrative to generate Haiku, which is naturally associated with the background story.

The genetic algorithms are inspired by the process of creating poems by human beings. Generally, people do not write a poem at one stroke. Instead, they start with an initial draft and go through several revisions to produce the final poem. Hrešková et al. \citep{hrevskova2017haiku} proposed an interactive genetic algorithm-based Haiku generation system. Users score each population of Haiku according to their subjective preferences, and the best Haiku in accordance with the user’s preference is obtained after multiple iterations. Yang et al. \citep{yang2016text} developed an interactive genetic algorithm-based Waka generation system which retrieved literary pieces related to the content and emotion of the user-provided texts from custom databases. After each generation, the system calculated the fitness of Waka using automatic and manual evaluation mechanisms, and produced the next population through a series of genetic operations.

With the rise of deep learning, poetry generation systems can better understand the semantic meaning of poetry through neural networks. Wu et al. \citep{wu2017haiku} studied Haiku generation with various RNN models and SeqGAN, and compared the perplexity values obtained from these models. Shao et al. \citep{shao2018traditional} used the LSTM language model to generate Haiku with specified keywords. Kaga et al. \citep{kaga2017learning} and Konishi et al. \citep{konishi2017generation} investigated how to generate Haiku preferred by the general public. Kaga et al. \citep{kaga2017learning} trained a LSTM language model with Haiku created by professional scholars, and then fine-tuned the model using Haiku created by non-professional scholars to make the generated poems easier to understand. Konishi et al. \citep{konishi2017generation} provided the generator and the discriminator of SeqGAN with different datasets, so that the discriminator could judge the generation with the values of the general public. Hirota et al. \citep{hirota2018haiku} adopted a method similar to that of Konishi et al. \citep{konishi2017generation}, aiming to avoid plagiarism when generating Haiku by Neural Probabilistic Language Model (NPLM). Masada et al. \citep{masada2018lda} trained a GRU language model with Waka poems segmented into bigrams, and scored the generated results using a LDA-based method. Their proposed scoring method could select generated poems with a wider variety of subsequences than those selected by RNN output probabilities.

In this paper, we conduct the study of Waka generation based on Transformer and/or VAE. On the one hand, neural networks are used to model the semantic relations in Waka, which could improve the model’s understanding of the semantic meaning. On the other hand, we consider both the objective and subjective methods to evaluate Waka poems generated by our models in a comprehensive way.

\section{Our approach}
\label{sec:3}
This section describes the method we used. We first formalize our task. Suppose $V$ denotes the vocabulary and $c\in V $ denotes a keyword specified by the user, our goal is to generate a Waka poem $x=(x_1,x_2,\cdots,x_T),x_i\in V$ of length $T$, which contains the given keyword $c$. According to the morae pattern, $x$ should be sequentially divisible into 5 phrases with fixed morae counts. That is to say, there exists exactly 4 subscripts $i,j,k,l$ subject to $1 \leq i<j<k<l<T$, making the following statements true:
\begin{equation}
	\sum_{t=1}^{i} s\left(x_{t}\right)=\sum_{t=j+1}^{k} s\left(x_{t}\right)=5
\end{equation}
\begin{equation}
	\sum_{t=i+1}^{j} s\left(x_{t}\right)=\sum_{t=k+1}^{l} s\left(x_{t}\right)=\sum_{t=l+1}^{T} s\left(x_{t}\right)=7
\end{equation}
where $s(w)$ denotes the morae count of a given word $w\in V$. As Fig.~\ref{fig:1} illustrates, a common separator Sep exists between the adjacent phrases, with the morae count defined as zero, i.e., $s(Sep)=0$.

Next, we introduce two Waka generation models based on the Transformer. The first model (TVAE) contains a single latent variable, while the second model (WakaVT) contains a sequence of latent variables.

\subsection{TVAE}
\label{sec:3.1}
Recently, the Transformer model and its variants \citep{tay2020efficient} have proven fruitful for a variety of NLP tasks. The original Transformer model \citep{vaswani2017attention} is an encoder-decoder framework designed for the sequence-to-sequence task. Each encoder layer consists of two sub-layers - the multi-head self-attention sublayer and the position-wise feed-forward network. Compared with the encoder layer, each decoder layer inserts an encoder-decoder attention sublayer between the above two sublayers to perform multi-head attention over the output of the encoder stack. Another difference between the encoder and the decoder is that the attention mechanisms of the former are non-causal, while those of the latter are causal, which is achieved through attention masks.

Self-attention mechanism is the core part of Transformer. In general, the attention mechanism learns the alignment between queries and keys through a compatibility function. For self-attention, a softmax function is applied to the output of the dot product between the queries and keys to obtain the alignment scores, which are then used to calculate the weighted sum of the values as output. This process is formulated as:
\begin{equation}
	\mathrm{Attention}(Q, K, V)=\mathrm{softmax}\left(\frac{Q K^{T}}{\sqrt{d_{k}}}\right) V
\end{equation}
where $Q$, $K$, and $V$ are the matrices of queries, keys, and values, respectively, and $d_k$ is the column count of $K$. The multi-head self-attention mechanism is then formulated as:
\begin{equation}
	\mathrm{MultiHead}(X)=\mathrm{concat}\left( \mathrm{head}_{1},\mathrm{head}_{2}, \ldots, \mathrm{head}_{h}\right) W^{o}
\end{equation}
\begin{equation}
	\mathrm{head}_{i}=\mathrm{Attention}\left(X W_{i}^{Q}, X W_{i}^{K}, X W_{i}^{V}\right)
\end{equation}
where $X$ is the matrix of the input sequence, and $W_i^Q,W_i^K,W_i^V,W^O$ are trainable parameters. In the self-attention mechanism, the dot product between the queries and keys drives the self-alignment process, where each token in the sequence learns to gather information from each other \citep{tay2020efficient}. On top of that, multi-head self-attention allows the model to focus on the information from different representation subspaces \citep{vaswani2017attention}. As we can see from the calculation, the self-attention mechanism fairly aligns any two positions in the input sequence (regardless of the distance between them), and thus can easily capture long-term dependencies.

\begin{figure}
	\centering
	\includegraphics[width=2.8in]{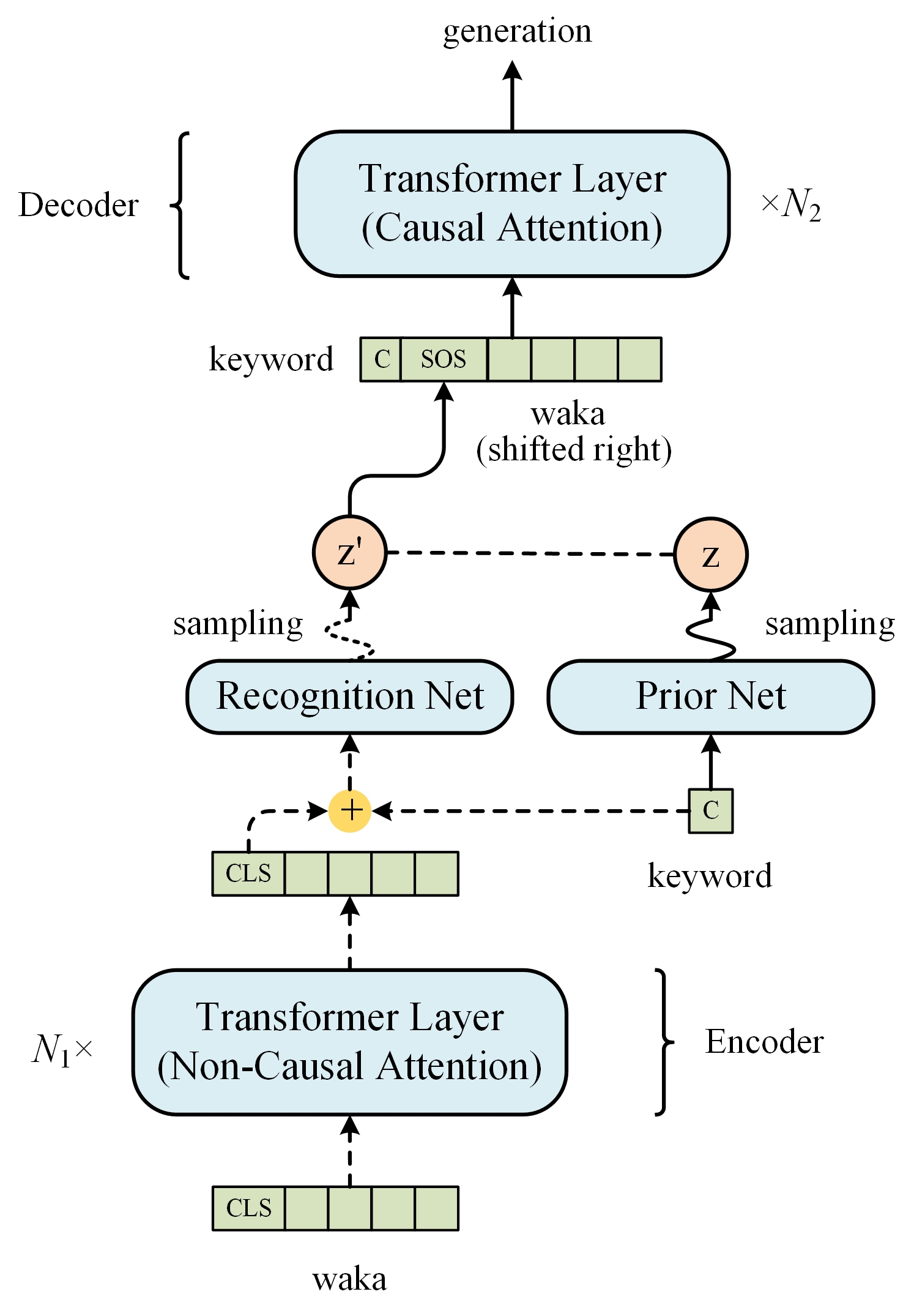}
	% figure caption is below the figure
	\caption{Architecture of TVAE. Dotted lines represent the connection that only appears in the training phase, and ``+" represents the concatenation operation.}
	\label{fig:2}       % Give a unique label
\end{figure}

\subsubsection{Model architecture}
\label{sec:3.1.1}
Based on the standard Transformer architecture, we construct the naive CVAE model (named TVAE) as a baseline. The overall structure of TVAE is shown in Fig.~\ref{fig:2}. We use precisely the same structure as the encoder layer of the original Transformer model on the encoder side. We leak the future information of the input sequence to the recognition network through the non-causal attention mechanism of the encoder layers. To encode the input sequence into a fixed-length vector, we follow the BERT model's approach \citep{devlin2019bert}, which adds a <CLS> token at the beginning of the input sequence and takes the output of the corresponding position as the encoded representation of the entire input. Both the recognition network and the prior network are Multilayer Perceptrons (MLPs) used to map inputs into the latent space. For the recognition network, the inputs consist of the output of the encoder and the embedding of the keyword. As a comparison, the prior network only takes the embedding of the keyword as input. The reparameterization trick \citep{kingma2014auto} is used here to solve the problem of non-differentiable sampling process of latent variables.

On the decoder side, we remove the encoder-decoder attention sublayer. We directly add the latent variable together with the embedding vector of <SOS> token, and then input the summation into the decoder. Inspired by the Ctrl model \citep{keskar2019ctrl}, we put the keyword at the beginning of the decoder input sequence as a control code to ensure that the generated poem contains this keyword. Thanks to the properties of the self-attention mechanism, the path length of any position in the sequence to the keyword is 1, thus enabling the keyword to have direct control over the content of the whole poem. Finally, the decoder generates a poem based on the latent variable and the input keyword. Note that the decoder layers contain causal attention since this allows the decoder to generate texts in an autoregressive manner.

Let $\theta$ be the parameters of the prior network and the decoder, and $\phi_r$ be the parameters of the encoder and the recognition network. Like the original CVAE model \citep{kihyuk2015learning}, the loss function of TVAE is defined as:
\begin{equation}
	\begin{aligned}
		L_{C V A E} &=-\mathbb{E}_{q_{\phi_{r}}(z \mid x, c)}\left[\log \frac{p_{\theta}(x, z \mid c)}{q_{\phi_{r}}(z \mid x, c)}\right] \\
		&=-\mathbb{E}_{q_{\phi_{r}}(z \mid x, c)}\left[\log p_{\theta}(x \mid z, c)\right]+D_{K L}\left(q_{\phi_{r}}(z \mid x, c) \| p_{\theta}(z \mid c)\right)
	\end{aligned}
\end{equation}
In this formula, $D_{KL}$ denotes the Kullback-Leibler divergence between the posterior and prior. We use the bag-of-word auxiliary loss \citep{zhao2017learning} and the KL cost annealing technique \citep{bowman2016generating} to alleviate the posterior collapse problem. The learning objective is finalized as:
\begin{equation}
	L=L_{CVAE}+\alpha L_{BOW}
\end{equation}
where $L_{BOW}$  denotes the bag-of-word auxiliary loss and $\alpha$ denotes a balancing coefficient.

\subsubsection{Morae constraint satisfaction using additive masks}
\label{sec:3.1.2}
Different words have different number of moraes, but the overall morae pattern of Waka is fixed, being 5, 7, 5, 7, 7. We design a mask-based method to ensure the generated poems follow the correct morae pattern. Let a 5-morae phrase or a 7-morae phrase in a Waka contain $n$ words, and the morae pattern of the phrase is expressed as a sequence $(a_1,a_2,\cdots,a_n)$, in which $a_i$ denotes the morae count of the \emph{i}-th word in the phrase. The total morae count $m=\sum_{i=1}^n{a_i}$ should satisfy $m=5$ or $m=7$, respectively. We define a sequence $(l_1,l_2,\cdots,l_n)$, in which $l_i=m-\sum_{k=1}^{i-1}{a_k}$ denotes the upper bound of the morae count of the \emph{i}-th word predicted by the decoder. Since the morae count of the \emph{i}-th word must not exceed $l_i$, we can mask out all words with the morae count larger than $l_i$ when the decoder predicts the \emph{i}-th word. Let the size of the dictionary be $D$, the morae count of the \emph{j}-th word $w_j$ in the dictionary be $s(w_j)$, and the logit and the predicted probability of the \emph{i}-th word be $o_i$ and $p_i=\mathrm{softmax}(o_i)$, respectively, then the mask $m_i=(m_{i,1},m_{i,2},\cdots,m_{i,D})$ at the current time step is defined as:
\begin{equation}
	m_{i, j}=\left\{\begin{aligned}
		0, & \text { if } \mathrm{s}\left(w_{j}\right) \leq l_{i} \\
		-\infty, & \text { if } \mathrm{s}\left(w_{j}\right)>l_{i}
	\end{aligned}\right.
\end{equation}
The output is then changed to $p_i^{\prime}=\mathrm{softmax}(o_i+m_i)$, where $p_i^{\prime}$ satisfies that for any $s(w_j)>l_i$, there is $p_{i,j}^{\prime}=0$. This mechanism masks out all words that may cause the morae count to exceed the limit when generating each word\footnote{Note that additive masks can also be used to avoid incomplete moraes. However, we have found that most generation could follow the correct pattern even if we don’t consider this problem.}.

In the training phase, additive masks can be calculated during data processing to speed up the training process. In the inference phase, they can be incrementally calculated. After generating a new word, the additive mask of the next word to be generated is calculated. Note that all special tokens (except for the generic unknown word token <UNK>) are specified with zero morae. For simplicity, the morae count of <UNK> is specified as an integer larger than 7 to ensure that no out-of-vocabulary (OOV) words appear in the generated results.

\subsection{WakaVT}
\label{sec:3.2}
TVAE uses a single latent variable that obeys Gaussian distribution, which has limited ability to model the diversity of the input data. In contrast, using a sequence of latent variables for text modeling can improve the diversity of the generated samples \citep{du2018variational,lin2020variational,schulz2018stochastic}. Inspired by this, we propose WakaVT, a Waka generation model based on the sequential variational Transformer. A sequence of latent variables is introduced to capture the variability of each position into a respective uni-modal distribution. In addition, WakaVT uses the same methods for keyword and morae constraint satisfaction as TVAE.

\subsubsection{Model architecture}
\label{sec:3.2.1}
\begin{figure}
	\centering
	\includegraphics[width=4.5in]{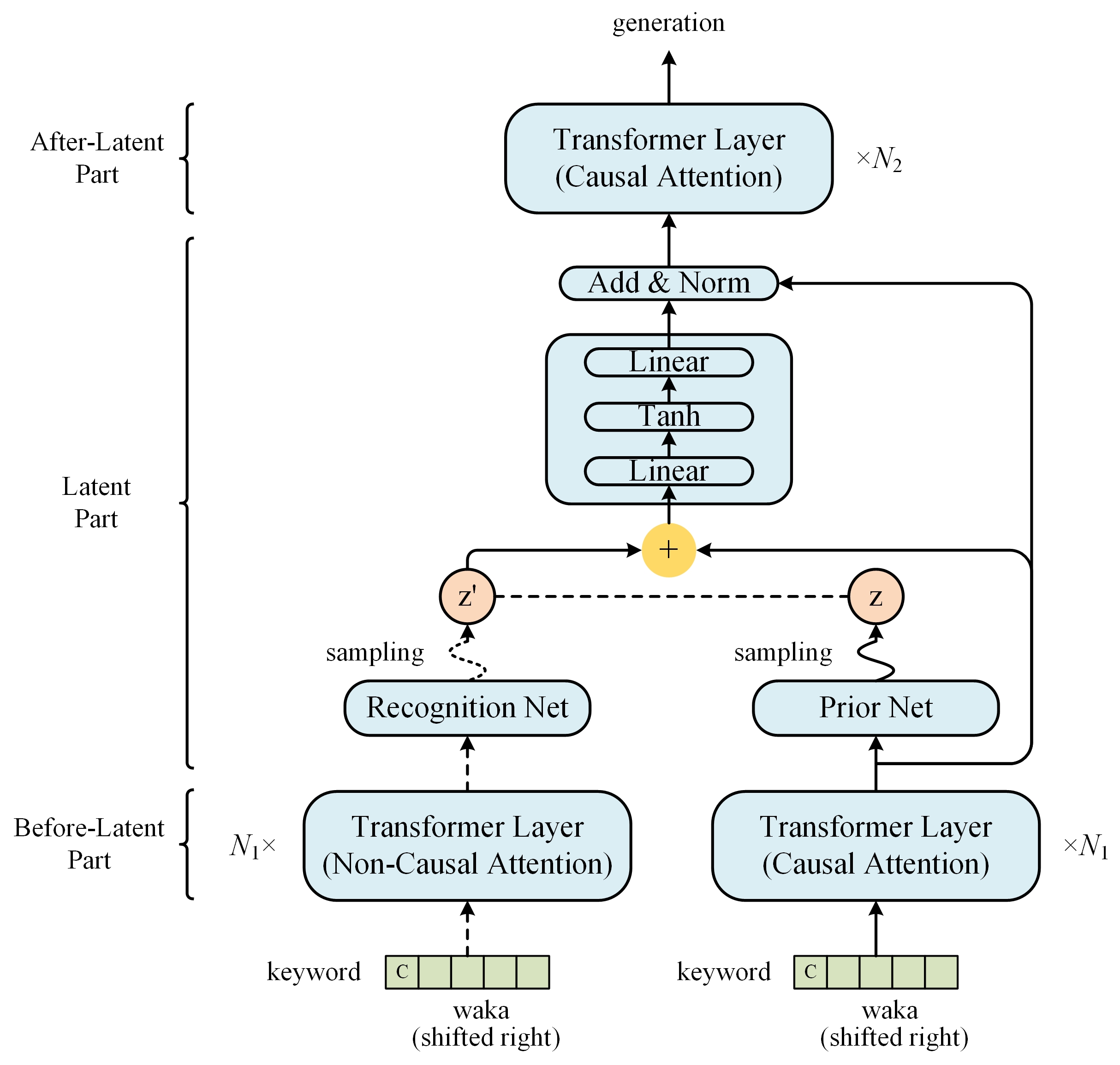}
	% figure caption is below the figure
	\caption{The architecture of WakaVT. Dashed lines represent connections that occur only during the training phase, and ``+" represents the concatenation operation.}
	\label{fig:3}       % Give a unique label
\end{figure}

The structure of WakaVT, as shown in Fig.~\ref{fig:3}, can be divided into three parts: the before-latent part, the latent part, and the after-latent part.

The before-latent part follows tightly after the input layer and is used to encode the input sequence and pass the observed information to the latent part. As in Fig.~\ref{fig:3}, there are 2 separate stacks of Transformer layers in this part. On the left side, a stack of $N_1$ layers adopts non-causal attention mechanism to allow the positions to attend to each other, which is similar to the encoder of TVAE. On the right side, another stack of $N_1$ layers contains causal attention to prevent the positions from attending to the subsequent positions. Similar to the decoder of TVAE, this allows the model to predict each token depending only on the known outputs when generating a text.

The latent part is the core of WakaVT, comprising the recognition network, the prior network, and the fusion network. The recognition network and the prior network are MLPs, which parameterize the conditional distributions of the corresponding latent variables $z^{\prime}$ and $z$. Assuming that the inputs of the recognition network and the prior network are $o^{(r)}$ and $o^{(p)}$, respectively, the posterior and prior at time step $t$ are approximated by:
\begin{equation}
	q_{\phi_{r}}\left(z_{t}^{\prime} \mid x_{1: T}, c\right)=\mathcal{N}\left(\mu_{t}^{\prime}, \sigma_{t}^{\prime 2}\right)
\end{equation}
\begin{equation}
	\left[\mu_{t}^{\prime}, \log \sigma_{t}^{\prime 2}\right]=\operatorname{MLP}\left(o_{t}^{(r)}\right)
\end{equation}
\begin{equation}
	q_{\phi_{p}}\left(z_{t} \mid x_{1:t-1}, c\right)=\mathcal{N}\left(\mu_{t}, \sigma_{t}^{2}\right)
\end{equation}
\begin{equation}
	\left[\mu_{t}, \log \sigma_{t}^{2}\right]=\operatorname{MLP}\left(o_{t}^{(p)}\right)
\end{equation}

The fusion network is among the recognition network, the prior network, and the after-latent part. It’s used to integrate the latent information represented by $z^{\prime}$ (in the training phase) or $z$ (in the inference phase), and the prior observed information represented by $o^{(p)}$, into a single representation. Notably, $z^{\prime}$ (or $z$) and $o^{(p)}$ are in different forms. The former is a randomly observed point in the latent space, while the latter is a deterministic text representation encoded by Transformer layers. Therefore, we introduce a fusion mechanism to merge them. The fusion process is calculated as:
\begin{equation}
	m_{t}=V \tanh \left(W z_{t}^{\prime}+U o_{t}^{(p)}\right)
\end{equation}
where $V$, $W$ and $U$ are trainable parameters. Afterwards, residual connection and residual dropout \citep{vaswani2017attention} are applied to make it easier for optimization. The final output of the latent-part is obtained as:
\begin{equation}
	o_{t}^{(m)}=\operatorname { LayerNorm }\left(o_{t}^{(p)}+\operatorname { Dropout }\left(m_{t}\right)\right)
\end{equation}

The after-latent part consists of a stack of $N_2$ Transformer layers and an output layer (not shown in Fig.~\ref{fig:3}). Its functionality is to predict the probability distribution of each token according to the latent information from the latent-part, the prior observed information from the before-latent part, and the input keyword from the user. During training, the decoding process of WakaVT is formulated as:
\begin{equation}
	p_{\theta}\left(x \mid z^{\prime}, c\right)=\prod_{t} p_{\theta}\left(x_{t} \mid x_{1: t-1}, z_{1:t}^{\prime}, c\right)
\end{equation}
where $\theta$ denotes trainable parameters of all parts of the model. During inference, since the recognition network is removed, we perform the decoding process using $z_{1:t}$ instead of $z_{1:t}^{\prime}$.

\subsubsection{Fused Multilevel Self Attention Mechanism}
\label{sec:3.2.2}
Waka has a natural hierarchical structure of four levels as shown in Fig.~\ref{fig:1}. There exist semantic connections and transitions among the phrases and the sentences. We take the Waka in Fig.~\ref{fig:1} as an example\footnote{Selected from the collection of \emph{Bunpo Hyakushu}, edited by \emph{Emperor Go-Uta}.}. The words むなしき(hollow), そら(sky), and かぜ(wind) in Kami-no-Ku are all about scenery and in close relation to each other. Similarly, the words ひと(people) and わすれ(forget) in Shimo-no-Ku are used to express the poet’s inner feelings and are also well correlated. However, there is a content transition from the scenery description to the psychological description, and the relations between the words in Kami-no-Ku and those in Shimo-no-Ku would not be so clear if we ignored the context. Therefore, semantic relations at different hierarchical levels may vary a lot in a Waka poem, and the relationship among the words more distant from each other is more difficult to understand.

\begin{figure}
	\centering
	\includegraphics[width=2.8in]{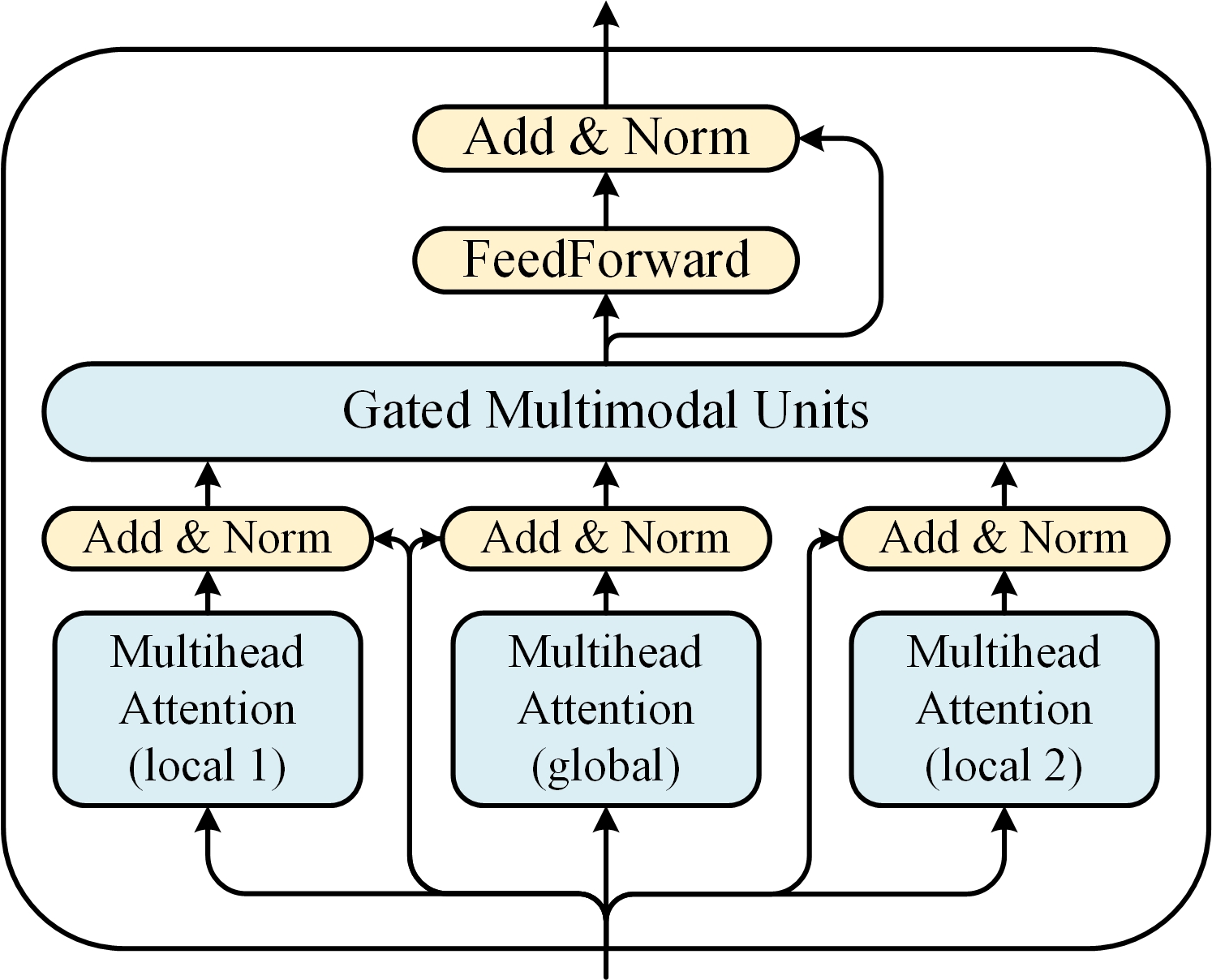}
	% figure caption is below the figure
	\caption{FMSA-based Transformer Layer}
	\label{fig:4}       % Give a unique label
\end{figure}

The original self-attention mechanism directly models the relationship between any two positions in the input sequence, unaware of the natural hierarchical feature of Waka. If such feature can be fully utilized, it will be easier for the model to understand the semantic relations at different levels. Therefore, we present the Fused Multilevel Self Attention (FMSA), which gathers features at different levels through different multi-head attention mechanisms and integrates them through a fusion mechanism as in Fig.~\ref{fig:4}. We incorporate FMSA into all Transformer layers of WakaVT.

\begin{figure*}
	\centering
	\subfigure[phrase level]{\includegraphics[width=1.5in]{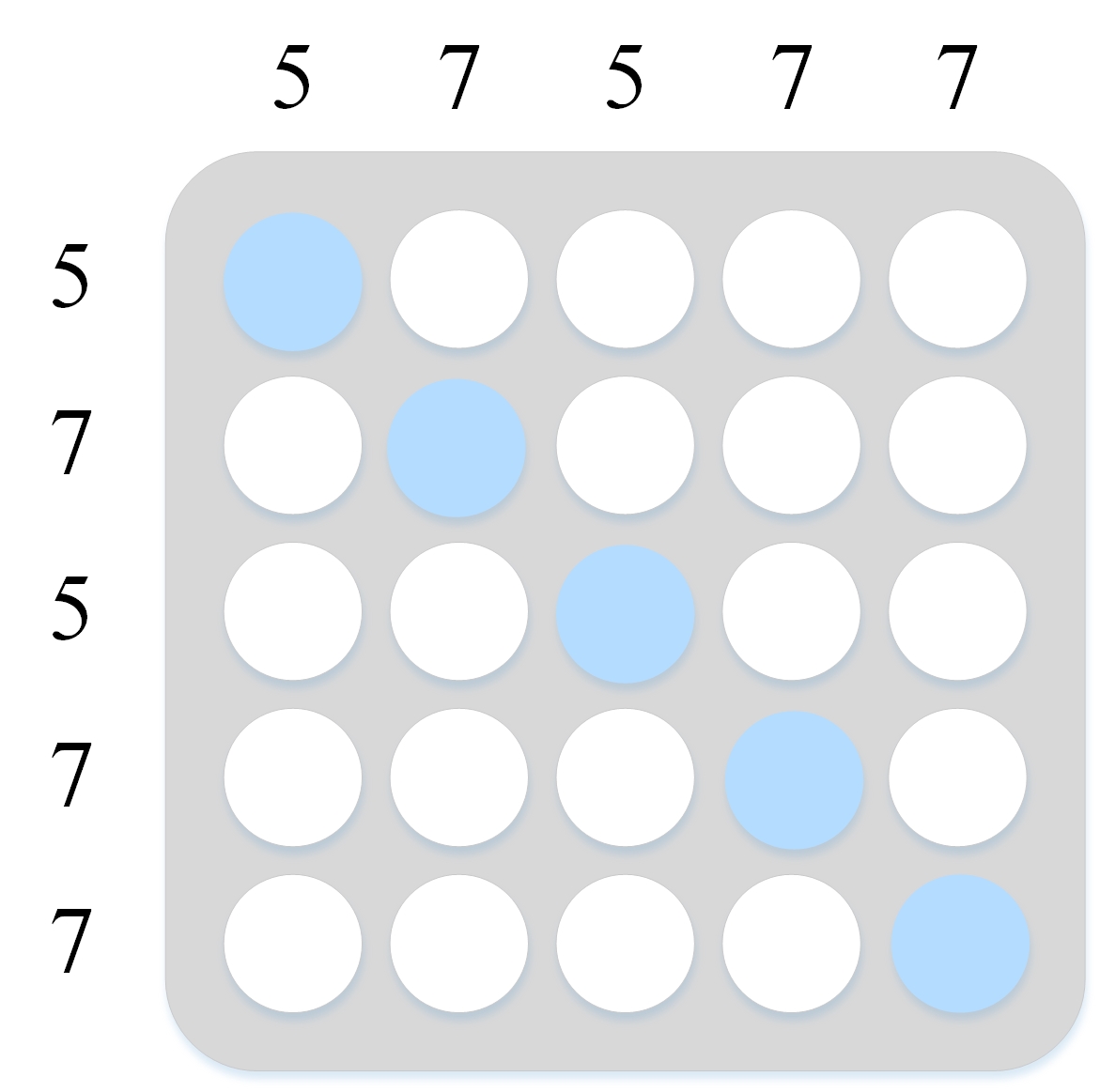} \label{fig:5a}}~~
	\subfigure[sentence level]{\includegraphics[width=1.5in]{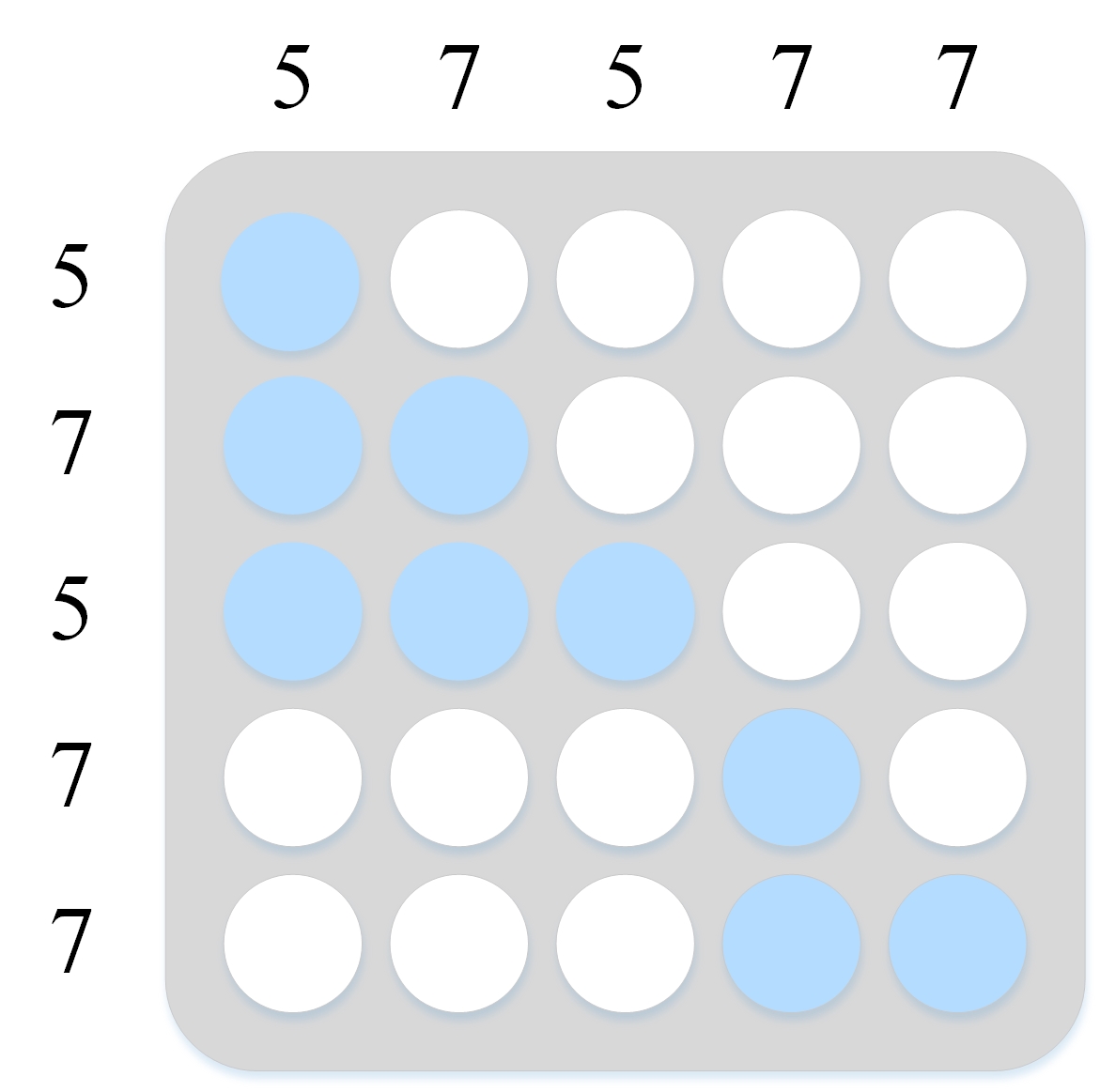} \label{fig:5b}}~~
	\subfigure[poem level]{\includegraphics[width=1.5in]{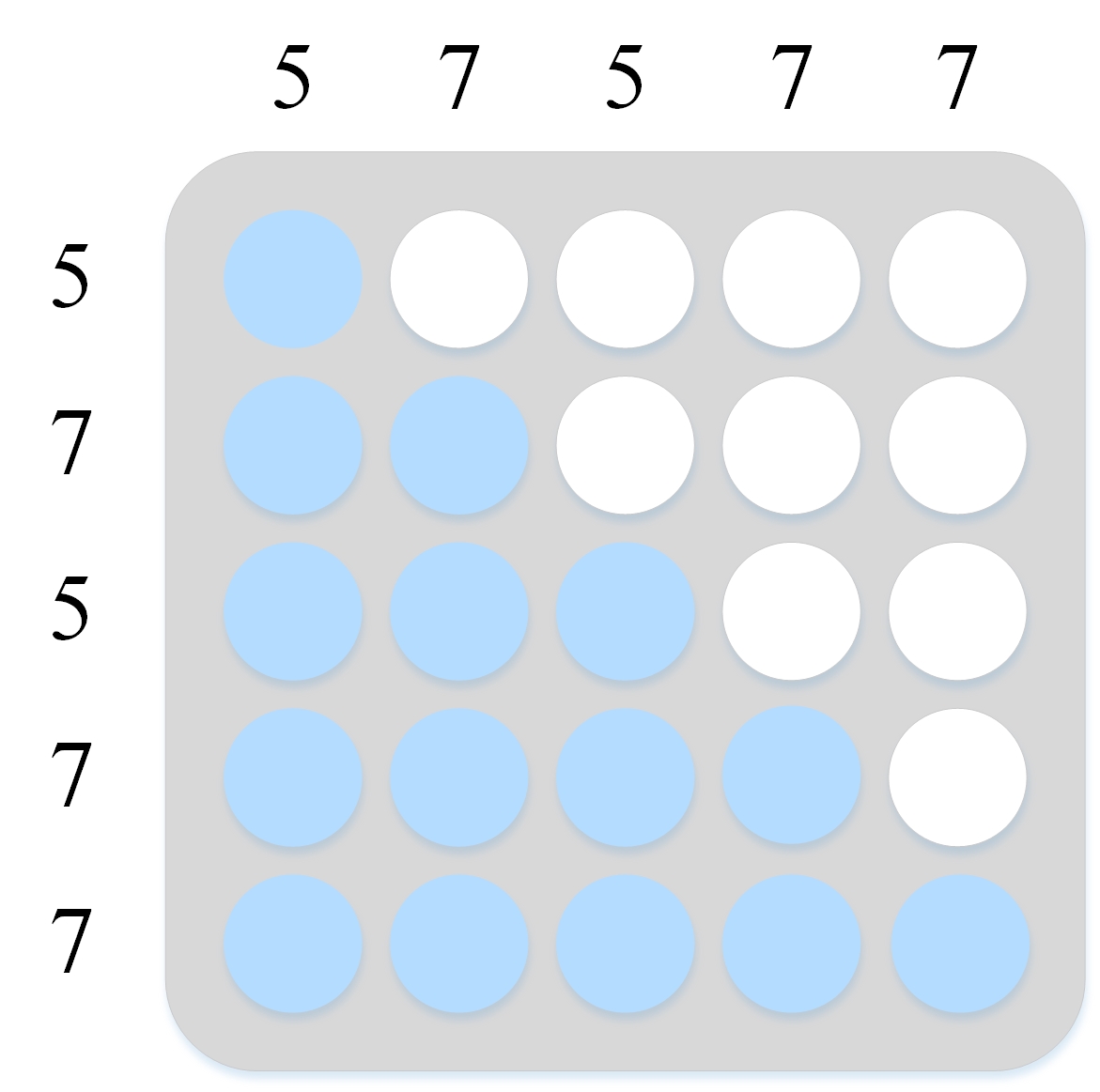} \label{fig:5c}}
	\caption{Attention Masks}
	\label{fig:5}
\end{figure*}

FMSA consists of three multi-head attention sublayers and a fusion unit. The global multi-head attention sublayer is applied to capture the global dependencies in Waka, as is the standard multi-head attention mechanism, connecting all pairs of the positions in the input sequence. Two local multi-head attention sublayers are introduced for the 5/7-morae phrase pattern and the sentence structure of Kami-no-Ku and Shimo-no-Ku, respectively. They restrict the self-alignment process in a local range to specially attend to the relations within phrases and sentences. Local attention mechanisms are implemented through attention masks. Fig.~\ref{fig:5} illustrates the format of the causal attention mask of each attention sublayer.

Supposing that after the residual connection and the layer normalization, the outputs of the three attention sublayers are $o^{f_1 },o^{f_2},o^{f_3}$, respectively (in no specific order), the fusion process is fulfilled through Gated Multimodal Unit (GMU) \citep{john2017gated}, which is calculated as:
\begin{equation}
	h_{t}^{f_{i}}=\tanh \left(W_{f_{i}} o_{t}^{f_{i}}\right)
\end{equation}
\begin{equation}
	z_{t}^{f_{i}}=\operatorname{sigmoid}\left(W_{z^{f_{i}}}\left[o_{t}^{f_{1}}, o_{t}^{f_{2}}, o_{t}^{f_{3}}\right]\right)
\end{equation}
\begin{equation}
	o_{t}^{\hat{f}}=\sum_{i=1}^{3} z_{t}^{f_{i}} * h_{t}^{f_{i}}
\end{equation}
\begin{equation}
	o^{\hat{f}}=\left(o_{1}^{\hat{f}}, o_{2}^{\hat{f}}, \ldots, o_{T}^{\hat{f}}\right)
\end{equation}
with $W_{f_i}$ and $W_{z^{f_i}}$ as trainable parameters and $o^{\hat{f}}$ as the output of GMU.

GMU first transforms the inputs from different modalities into feature vectors through nonlinear transformations, and then controls the contribution of the calculated features to the overall output through gating strategies using sigmoid function. Intuitively, GMU guides FMSA to explore how to integrate the most essential features from different hierarchical levels into a single vector. In this way, the Transformer layers with FMSA can achieve fine-grained representations of hierarchical features of the input sequence.

\subsubsection{Learning}
\label{sec:3.2.3}
Given a keyword $c$ and a sample $x=x_{1:T}$ of length $T$, the loss function of WakaVT can be written as:
\begin{equation}
	\begin{aligned}
		L_{C V A E}=&-\mathbb{E}_{q_{\phi_{r}}\left(z_{1: T} \mid x_{1: T}, c\right)}\left[\sum_{t} \log \frac{p_{\theta}\left(x_{t} \mid x_{1: t-1}, z_{1: t}, c\right) p_{\theta}\left(z_{t} \mid x_{1: t-1}, c\right)}{q_{\phi_{r}}\left(z_{t} \mid x_{1: T}, c\right)}\right] \\
		=&-\sum_{t} \mathbb{E}_{q_{\phi_{r}}\left(z_{1: t} \mid x_{1: T}, c\right)}\left[\log p_{\theta}\left(x_{t} \mid x_{1: t-1}, z_{1: t}, c\right)\right] \\
		&+\sum_{t} D_{K L}\left(q_{\phi_{r}}\left(z_{t} \mid x_{1: T}, c\right) \| q_{\phi_{p}}\left(z_{t} \mid x_{1: t-1}, c\right)\right)
	\end{aligned}
\end{equation}
where $D_{KL}$ is the Kullback-Leibler divergence between the posterior and prior.

To alleviate the posterior collapse problem, we introduce the SBOW auxiliary loss \citep{du2018variational}, which forces the latent variables to capture more useful information from the input sequence. It is defined as:
\begin{equation}
	p_{\xi}\left(x_{t: t+l-1} \mid z_{t}, x_{1:t-1}, c\right)=\operatorname{MLP}\left(z_{t}, o_{t}^{(p)}\right)
\end{equation}
\begin{equation}
	L_{SBOW}=-\sum_{t} \mathbb{E}_{q_{\phi_{r}}\left(z_{t} \mid x_{1: T}, c\right)}\left[\log p_{\xi}\left(x_{t: t+l-1} \mid x_{1: t-1}, z_{t}, c\right)\right]
\end{equation}
where $l$ is the truncated length, which simply means that only the continuous $l$ tokens started from $x_t$ are considered when making the calculation at current time step $t$. Ultimately, the total loss of WakaVT is the weighted sum of $L_{CVAE}$ and $L_{SBOW}$:
\begin{equation}
	L=L_{CVAE}+\alpha L_{SBOW}
\end{equation}

\section{Experiments}
\label{sec:4}

\subsection{Data}
\label{sec:4.1}
We build a large and available Waka dataset containing anthologies from the 8th to 16th centuries AD, including \emph{Manyoshu}, \emph{Nijuichidaishu}, \emph{Uta-awase} and a variety of private collections. All data are collected from the \emph{International Research Center for Japanese Studies} database\footnote{http://db.nichibun.ac.jp/pc1/ja/category/waka.html}. After data cleansing, we obtained a total of 171801 Waka poems, each of which is written in one continuous line using historical kana orthography. For the sake of availability, the voiced sounds need to be recognized and marked with dakuten diacritic marks, and the kana characters should be segmented into words to build the vocabulary. Therefore, we first train a sequence labeling model based on Conditional Random Fields (CRF) \citep{charles2012an} using 20 thousand Waka poems with properly labeled voiced sounds, and then use it to automatically annotate our dataset. Next, we applied Web茶まめ\footnote{https://chamame.ninjal.ac.jp/}(\emph{Web Chamame}), a morphological analysis tool built by the \emph{National Institute for Japanese Language and Linguistics}, to perform word segmentation. TextRank algorithm \citep{yang2018generating} is adopted to extract a keyword for each Waka. Finally, 10,000 poems are randomly selected from the dataset as the validation set, 5,000 as the test set, and the remaining 156,801 as the training set.

\subsection{Baselines}
\label{sec:4.2}
For Waka generation with given keywords, we design and implement three baselines as below. To ensure the generated Waka strictly conforms to the morae constraint, the additive mask-based method proposed in Sect.~\ref{sec:3.1.2} is applied to each model.

\textbf{RNN-VAD:} A GRU-based conditional variational autoregressive decoder. A sequence of latent variables subject to a multimodal distribution is used to model the high variability of texts. It is adapted from the model proposed in \citep{du2018variational}. We retained the forward RNN and the backward RNN, removed the encoder and the associated attention module, and added a linear layer for each RNN to convert the embedding vector of the input keyword into the initial hidden states.

\textbf{TLM:} A Transformer-based conditional language model. Texts with controllable content can be generated using various control codes \citep{keskar2019ctrl}.

\textbf{TVAE:} A Transformer-based CVAE model, as discussed in Sect.~\ref{sec:3.1}. Unlike WakaVT, TVAE models data using the unimodal distribution of a single latent variable.

FMSA is proposed based on the self-attention mechanism and is independent of specific model architectures. We applied FMSA to TLM, TVAE and WakaVT to investigate its effect on each model.

\subsection{Model settings}
\label{sec:4.3}
The 128-dimensional word embeddings pretrained by Fasttext \citep{bojanowski2017enriching} were adopted to initialize each model's embedding layer, with a dictionary size of 6649. For all Transformer-based models, the multi-head attention sublayer was made up of 4 heads and the feedforward network contained an inner layer with 512 units. The TLM model consisted of a stack of 4 Transformer layers, and the same setting was adopted for the counterparts of TVAE and WakaVT. Specifically, for TVAE, the corresponding parameters $N_1$, $N_2$ (see Fig.~\ref{fig:2}) were set as $N_1=N_2=4$, and for WakaVT (Fig.~\ref{fig:3}), $N_1=N_2=2$. Both the forward and the backward GRU of RNN-VAD contained 3 layers with a hidden size of 256, and layer normalization was applied to each GRU cell as in \citep{ba2016layer}. The size of the latent variables was set to 128 everywhere. In the training stage, the models were all trained by Adam optimizer \citep{diederik2015adam}, with a learning rate of 0.0001 and a mini-batch size of 32. Linear scheduling of KLD loss was applied to the training process of all VAE models, combined with appropriate auxiliary loss to alleviate the posterior collapse. Specifically, BOW loss was adopted for TVAE, while SBOW loss with a truncated length of 5 was adopted for RNN-VAD and WakaVT. In the inference stage, we used beam search algorithm with the beam width set to 20 for all models.

\subsection{Evaluation design}
\label{sec:4.4}
It is notoriously difficult to judge the quality of poems generated by computers. We conduct both objective and subjective evaluation to comprehensively evaluate the poems generated by each model.

\subsubsection{Objective evaluation}
\label{sec:4.4.1}
We employ three evaluation metrics in our experiments. They are defined as follows:

\textbf{PPL \& KLD:} We calculated the reconstruction perplexity (PPL) and the Kullback-Leibler divergence (KLD) between the posterior and prior for each model on the test set. A well-trained model should achieve a low PPL value and a small but non-trivial KLD value.

\textbf{Novelty:} The novelty is aimed to evaluate to what extent the generated Waka differs from the training set: whether the model simply copies the pieces from the training set or generates new pieces itself. Given a training set $C$ and a generation set $S$, $\mathrm{Nov}_w$ is defined as:
\begin{equation}
	\operatorname{Nov}_{w}(S)=\frac{1}{|S|} \sum_{i=1}^{|S|} \operatorname{Nov}_{w}\left(S_{i}\right)
\end{equation}
\begin{equation}
	\operatorname{Nov}_{w}\left(S_{i}\right)=1-\max \left\{\operatorname{Dice}\left(S_{i}, C_{j}\right)\right\}_{j=1}^{j=|C|}
\end{equation}
with $S_i$ as a poem in $S$, $C_j$ as a poem in $C$, and $\operatorname{Dice}(S_i,C_j)$ as the Sørensen-Dice coefficient\footnote{https://en.wikipedia.org/wiki/S\%C3\%B8rensen\%E2\%80\%93Dice\_{}coefficient} between $S_i$ and $C_j$. Intuitively, $\mathrm{Nov}_w$ only measures the novelty at the word level, thus we define a metric at the phrase level to make the evaluation more comprehensive. Given a set of Waka $A$, $\operatorname{Phr}_5(A)$ and $\operatorname{Phr}_7(A)$ as the set of the corresponding 5-morae phrases and 7-morae phrases, respectively, $\operatorname{Nov}_{s,n}$ is defined as:
\begin{equation}
	\operatorname{Nov}_{s, n}(S)=\frac{\left|\operatorname{Phr}_{\mathrm{n}}(S)-\operatorname{Phr}_{\mathrm{n}}(C)\right|}{\left|\mathrm{Phr}_{\mathrm{n}}(S)\right|}, n=5 \text { or } 7
\end{equation}
We use 1000 keywords to generate 1000 Waka poems (one for each keyword) for each model to calculate the above metrics.

\textbf{Diversity:} The diversity is designed to measure how different the generated poems are with different specified inputs. A low diversity metric signals that the model always generates similar results and lacks creative potential. Similar to the novelty metrics, we define a metric at the word level as $\mathrm{Div}_w$ and another at the phrase level as $\operatorname{Div}_{s,n}$:
\begin{equation}
	\operatorname{Div}_{w}(S)=\frac{1}{|S|} \sum_{i=1}^{|S|} \operatorname{Div}_{w}\left(S_{i}\right)
\end{equation}
\begin{equation}
	\operatorname{Div}_{w}\left(S_{i}\right)=1-\max \left\{\operatorname{Dice}\left(S_{i}, S_{j}\right)\right\}_{j=1}^{j=|S|, j \neq i}
\end{equation}
\begin{equation}
	\operatorname{Div}_{s, n}(S)=\frac{\left|\operatorname{Phr}_{\mathrm{n}}(S)\right|}{\sum_{p \in \operatorname{Phr}_{\mathrm{n}}(s)} \operatorname{Count}_{S}(p)}, n=5 \text { or } 7
\end{equation}
where $\operatorname{Count}_{S}(p)$ denotes the number of times the phrase $p$ appears in $S$. The novelty metric and diversity metric are calculated using the same generation set $S$.

\subsubsection{Subjective evaluation}
\label{sec:4.4.2}

%\begin{table}
%	% table caption is above the table
%	\caption{Subjective evaluation criteria}
%	\label{tab:1}       % Give a unique label
%	% For LaTeX tables use
%	\begin{tabu} to \hsize {XX[5,l]}
%		\hline
%		Fluency & Whether a poem is grammatically and syntactically satisfied. \\ \hline
%		Coherence & Whether a poem is semantically coherent among the phrases and between Kami-no-ku and Shimo-no-ku. \\ \hline
%		Meaningfulness & Whether a poem conveys an artistic conception with an optional use of figures of speech, such as kakekotoba, engo, and makurakotoba. \\ \hline
%		Overall & Mean value of the above three metrics. \\
%		\hline
%	\end{tabu}
%\end{table}
We subjectively evaluate the linguistic quality of Waka from three aspects: fluency, coherence, and meaningfulness. The details are shown in Table ~\ref{tab:1}. In the experiment, it was found that the additive mask-based method proposed in Sect.~\ref{sec:3.1.2} was sufficient to ensure that most generated poems met the morae constraint\footnote{Among the generated poems used for objective evaluation, the proportion of morae constrained poems generated by TLM, TVAE, WakaVT, and RNN-VAD were 92.7\%, 98.2\%, 99.5\%, and 100\%, respectively.}. Therefore, we do not design a particular evaluation method for the correctness of formats. For the sake of fairness, the 100 most frequent keywords were selected, and each model generated a poem for each keyword. We invited 3 experts major in classical Japanese poetry to conduct a blind review of the generated poems on a scale from 1 (very poor) to 5 (very good). The scores of the poems are averaged as the final score for each model.

\begin{table}[htbp]
	\caption{Subjective evaluation criteria}
	\label{tab:1}
	\begin{tabularx}{\linewidth}{Xp{.7\linewidth}}
		\hline\noalign{\smallskip}
		Metric & Detail   \\
		\noalign{\smallskip}
		\hline\noalign{\smallskip}
		Fluency & Whether a poem is grammatically and syntactically satisfied. \\
		Coherence & Whether a poem is semantically coherent among the phrases and between Kami-no-ku and Shimo-no-ku.   \\
		Meaningfulness & Whether a poem conveys an artistic conception with an optional use of figures of speech, such as kakekotoba, engo, and makurakotoba.   \\
		Overall & Mean value of the above three metrics.   \\
		\noalign{\smallskip}\hline
	\end{tabularx}
\end{table}

\subsection{Quantitative analysis}
\label{sec:4.5}

\begin{table}[htbp]
	\caption{Objective evaluation results. ↓ indicates the lower the better, while ↑ indicates the higher the better.}
	\label{tab:2}
	\begin{tabularx}{\linewidth}{Xp{.065\linewidth}<{\centering}p{.065\linewidth}<{\centering}p{.065\linewidth}<{\centering}p{.065\linewidth}<{\centering}p{.065\linewidth}<{\centering}p{.065\linewidth}<{\centering}p{.065\linewidth}<{\centering}p{.065\linewidth}<{\centering}}
		\hline\noalign{\smallskip}
		\multirow{2}{*}{Model} & \multirow{2}{*}{PPL↓} & \multirow{2}{*}{KLD↑} &
		\multicolumn{3}{c}{Novelty↑} & \multicolumn{3}{c}{Diversity↑} \\
		\cmidrule(lr){4-9}
		& & & $\mathrm{Nov}_w$ & $\mathrm{Nov}_{s,5}$ & $\mathrm{Nov}_{s,7}$ & $\mathrm{Div}_w$ & $\mathrm{Div}_{s,5}$ & $\mathrm{Div}_{s,7}$ \\
		\noalign{\smallskip}
		\hline\noalign{\smallskip}
		TLM & 15.14	& -	& 0.3909 & 0.0661 & 0.2104 & 0.2979 & 0.4326 & 0.5110 \\
		TLM+FMSA & 14.21 & - & 0.3776 & 0.0249 & 0.1153 & 0.2999 & 0.3801 & 0.4683 \\
		TVAE & 12.72 & 10.71 & 0.4161 & 0.0767 & 0.3675 & 0.4394 & 0.3651 & 0.5903 \\
		TVAE+FMSA & 12.09 & 8.66 & 0.4027 & 0.0351 & 0.2258 & 0.4128 & 0.3287 & 0.5237 \\
		RNN-VAD & 8.09 & 18.86 & 0.4201 & \textbf{0.1792} & 0.4962 & 0.4883 & 0.6055 & 0.8243 \\
		WakaVT & 6.98 & 20.18 & 0.4273 & 0.1482 & 0.4542 & 0.4978 & \textbf{0.6171} & 0.7966 \\
		WakaVT+FMSA & \textbf{5.60} & \textbf{26.70} & \textbf{0.4400} & 0.1446 & \textbf{0.4966} & \textbf{0.5182} & 0.6085 & \textbf{0.8310} \\
		\noalign{\smallskip}\hline
	\end{tabularx}
\end{table}

The objective evaluation results are shown in Table ~\ref{tab:2}. We can find that models with a sequence of latent variables (WakaVT, WakaVT+FMSA, and RNN-VAD) achieved smaller PPL values and larger KLD values than those with a single latent variable (TVAE and TVAE+FMSA). Additionally, TVAE and TVAE+FMSA got smaller PPL values than the models without latent variables (TLM and TLM+FMSA). This indicates that a single latent variable can improve the model’s capability to reconstruct the input sequence, and a sequence of latent variables may further boost the results by capturing more detailed semantic information. Compared with WakaVT and RNN-VAD, WakaVT+FMSA achieved the lowest PPL value and the highest KLD value, due to the high modeling capability of the Transformer architecture along with the auxiliary effect of the FMSA module.

It’s observed that models with a sequence of latent variables outperformed other models in terms of $\mathrm{Nov}_w$, $\mathrm{Nov}_{s,5}$, and $\mathrm{Nov}_{s,7}$, which confirms that latent variables can indeed improve the novelty of words and phrases. Compared with RNN-VAD, WakaVT+FMSA achieved larger $\mathrm{Nov}_w$, smaller $\mathrm{Nov}_{s,5}$, and similar $\mathrm{Nov}_{s,7}$. This demonstrates that WakaVT+FMSA has comparable capability to RNN-VAD in terms of novelty.

Through the Pearson Correlation Analysis, we find that $\mathrm{Nov}_w$ and $\mathrm{Div}_w$ show a strong positive correlation ($r=0.96$). The result still holds for the correlation between $\mathrm{Nov}_{s,5}$ and $\mathrm{Div}_{s,5}$ ($r=0.94$), and that between $\mathrm{Nov}_{s,7}$ and $\mathrm{Div}_{s,7}$ ($r=0.96$). Similar to the conclusion of the novelty evaluation, models with a sequence of latent variables outperformed other models in terms of $\mathrm{Div}_w$, $\mathrm{Div}_{s,5}$, and $\mathrm{Div}_{s,7}$. Among them, WakaVT+FMSA scored the highest in $\mathrm{Div}_w$ and $\mathrm{Div}_{s,7}$ and WakaVT scored the highest in $\mathrm{Div}_{s,5}$. This proves that a sequence of latent variables can improve the diversity at both the word level and the phrase level. Additionally, the combination of the Transformer and FMSA can further boost the results.

\begin{table}[htbp]
	\caption{Subjective evaluation results}
	\label{tab:3}
	\begin{tabularx}{\linewidth}{Xp{.17\linewidth}<{\centering}p{.17\linewidth}<{\centering}p{.17\linewidth}<{\centering}p{.17\linewidth}<{\centering}}
		\hline\noalign{\smallskip}
		Model & Fluency & Coherence & Meaningfulness & Overall \\
		\noalign{\smallskip}
		\hline\noalign{\smallskip}
		TLM & 4.30 & 3.66 & 2.93 & 3.63
		\\
		TLM+FMSA & 4.49 & 3.98 & 3.49 & 3.99
		\\
		TVAE & 4.36 & 3.69 & 3.16 & 3.74
		\\
		TVAE+FMSA & \textbf{4.57} & 3.94 & 3.55 & 4.02
		\\
		RNN-VAD & 4.38 & 3.83 & 3.37 & 3.86
		\\
		WakaVT & 4.36 & 3.84 & 3.38 & 3.86
		\\
		WakaVT+FMSA & 4.52 & \textbf{4.17} & \textbf{3.74} & \textbf{4.14}
		\\
		\noalign{\smallskip}\hline
	\end{tabularx}
\end{table}

Table ~\ref{tab:3} shows the results of the subjective evaluation. For each model, the scores of different metrics could be ranked in a decreasing order as fluency, coherence, meaningfulness. This indicates that for any model, it is relatively easy to form smooth sentences, rather difficult to ensure semantic coherence, and the greatest challenge is to express desired artistic conception. Significantly, models with FMSA module (TLM+FMSA, TVAE+FMSA, WakaVT+FMSA) obtained higher scores of all metrics than the other models, confirming the effect of FMSA in improving the linguistic quality of the generated poems. Among the models with FMSA module, WakaVT+FMSA scored the highest while TLM+FMSA and TVAE +FMSA obtained similar scores. Specifically, TVAE+FMSA scored the highest in fluency, while WakaVT+FMSA scored the second-best with small difference from TVAE+FMSA. In terms of coherence and meaningfulness, TLM+FMSA and TVAE+FMSA scored similarly, while WakaVT+FMSA had a significant advantage over them ($p<0.05$)\footnote{Unilateral T-test is used for significance test.}.

Overall, RNN-VAD performed similar to WakaVT+FMSA in terms of novelty and diversity, yet worse than WakaVT+FMSA in terms of linguistic quality. TLM+FMSA and TVAE+FMSA were not as good at linguistic quality as WakaVT+FMSA, and were even significantly inferior to WakaVT+FMSA in novelty and diversity. Therefore, we conclude that WakaVT+FMSA outperformed baseline models significantly in the overall quality of generated poems, including linguistic quality, novelty and diversity.

\subsection{Qualitative analysis}
\label{sec:4.6}

\subsubsection{FMSA visualizations}
\label{sec:4.6.1}

\begin{figure*}
	\centering
	\subfigure[FMSA local attention at the phrase level]{\includegraphics[width=2.8in]{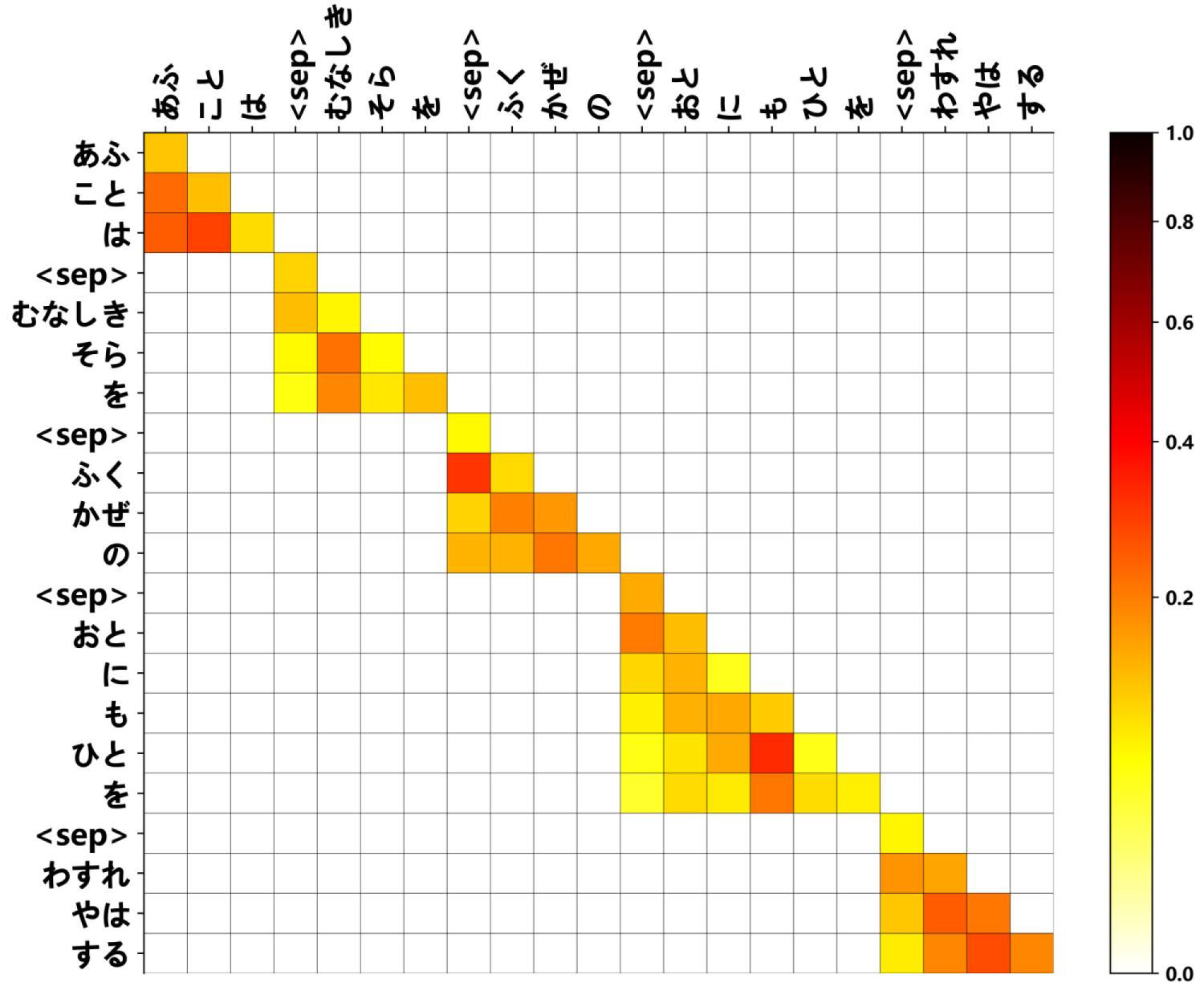} \label{fig:6a}}~~
	\subfigure[FMSA local attention at the sentence level]{\includegraphics[width=2.8in]{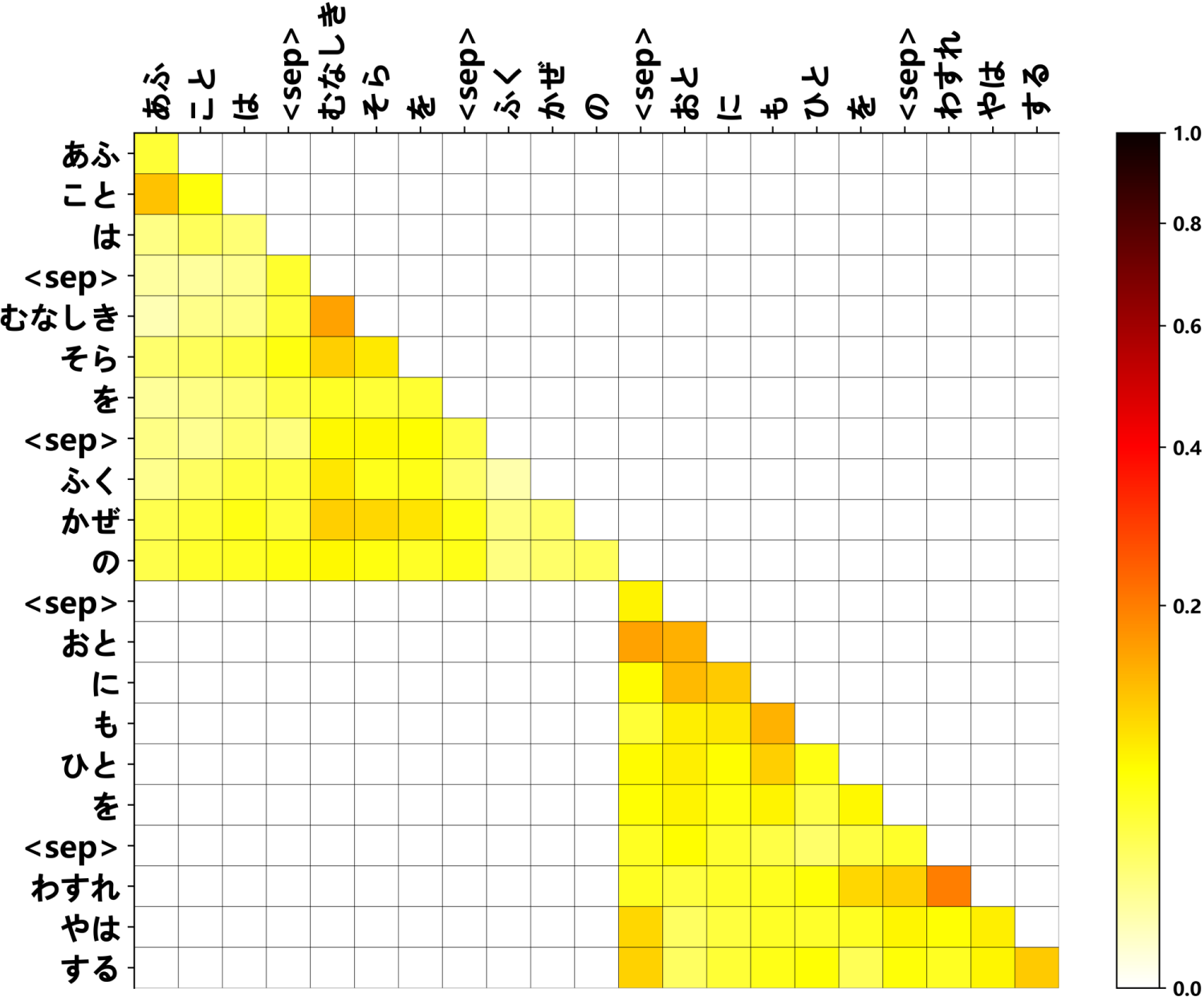} \label{fig:6b}}
	\subfigure[FMSA global attention]{\includegraphics[width=2.8in]{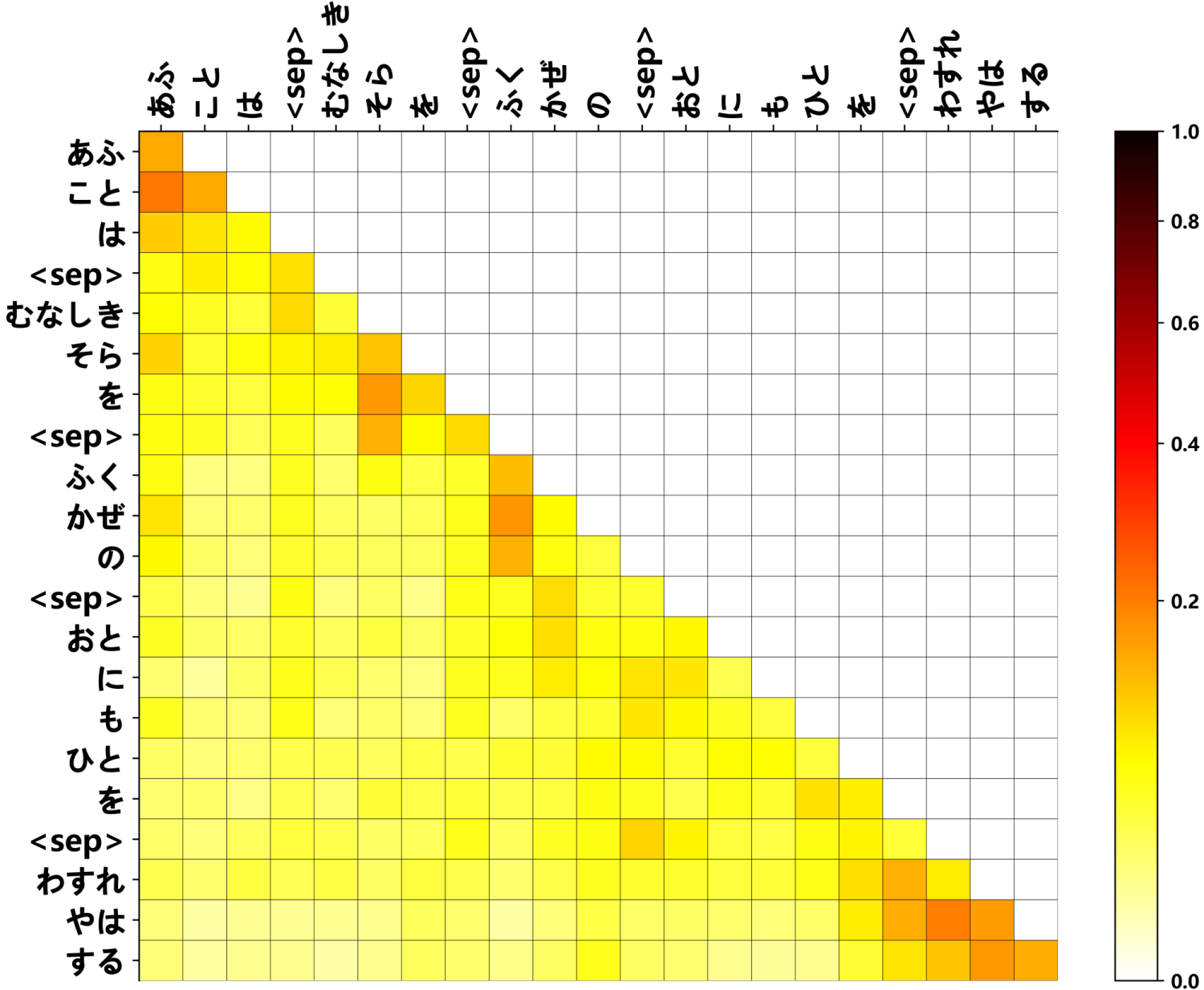} \label{fig:6c}}~~
	\subfigure[standard self-attention]{\includegraphics[width=2.8in]{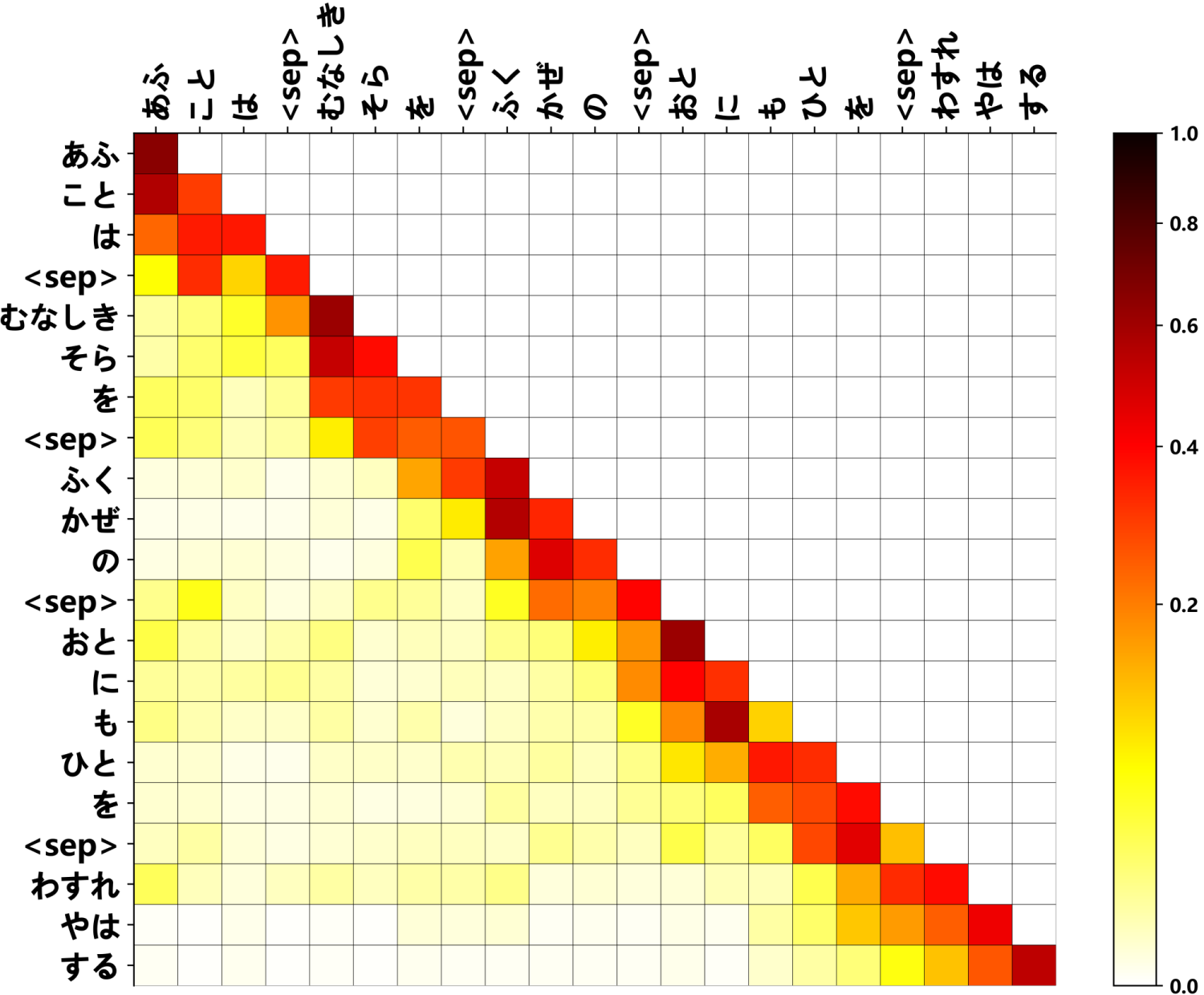} \label{fig:6d}}
	\caption{FMSA visualization results}
	\label{fig:6}
\end{figure*}

To analyze the mechanism of FMSA more intuitively, we input a Waka poem from the test set (same as the example discussed in Sect.~\ref{sec:3.2.2}) into WakaVT and WakaVT+FMSA, and conduct a visualization analysis of the alignment weights. The calculated PPL values of WakaVT and WakaVT+FMSA with this poem as input are 5.39 and 3.15, respectively. We visualized the alignment weights at the first layer of the after-latent part (averaging the weights of all 4 heads), with results shown in Fig.~\ref{fig:6}.

The visualization results of the self-attention mechanisms at different levels are shown in Fig.~\ref{fig:6a}, Fig.~\ref{fig:6b}, and Fig.~\ref{fig:6c}, respectively. It can be seen that FMSA local attention (Fig.~\ref{fig:6a}) at the phrase level only aligns the tokens within 5/7-morae phrases. For example, in the second phrase, the local attention attends to the dependency between むなしき(hollow) and そら(sky). FMSA local attention at the sentence level (Fig.~\ref{fig:6b}) aligns the tokens within Kami-no-ku and Shimo-no-ku. We can see that the model pays extra attention to the lexical dependencies among different phrases in a sentence, and the dependencies within the phrases are weakened (the color becomes lighter). For example, the alignment weight between むなしき(hollow) and そら(sky) has decreased, meanwhile the model puts a special emphasis on the connection between かぜ(wind) and the above two words. The FMSA global attention (Fig.~\ref{fig:6c}) aligns all the words in the Waka. It is observed that much attention is paid to adjacent words, but the dependencies within Kami-no-ku and Shimo-no-ku, as well as those between the sentences, are also taken into consideration. For example, the connections among あふ(meeting), そら(sky), and かぜ(wind) in the Kami-no-ku (the wind in the sky is a metaphor for lovers meeting), and the connection between かぜ(wind) in the Kami-no-ku and おと(sound) in the Shimo-no-ku are marked in deeper colors. The visualization results of the standard self-attention are shown in Fig.~\ref{fig:6d}. Compared with FMSA global attention, it pays little attention to the connections among the words located far from each other, but too much attention to local lexical connections.

The following observations are drawn from the above results. Firstly, the standard self-attention mechanism models the lexical connections in Waka from only a single view, while FMSA is able to do it from 3 views through different self-attention modules. As a result, FMSA can deal with the lexical connections at different hierarchical levels, effectively leveraging hierarchical features of Waka. Secondly, the standard self-attention mechanism may tend to model local lexical connections at the phrase level, ignoring certain important connections in a wider range, which is more difficult to capture. On the contrary, for FMSA, the phrase-level dependency modeling is completed by the corresponding local attention sublayer, leaving the other two attention sublayers to capture longer dependencies.

\subsubsection{Case study}
\label{sec:4.6.2}

\begin{table}[htbp]
	\caption{Generated results of each model with keyword はる(spring). Each poem is splitted into two lines, i.e. Kami-no-ku and Shimo-no-ku. Repeated words/phrases are underlined.}
	\label{tab:4}
	\begin{tabularx}{\linewidth}{Xp{.7\linewidth}<{\centering}}
		\hline\noalign{\smallskip}
		Model & Generated Waka  \\
		\noalign{\smallskip}\hline\noalign{\smallskip}
		\multirow{4}{*}{TLM+FMSA} & はるのきる−\underline{かすみのころも}−うすければ \\
		& \underline{かすみのころも}−ほころびにけり \\
		& Spring comes in rosy cloud dress. \\
		& Too thin the cloth is, the dress splits. \\
		\noalign{\smallskip}\hline\noalign{\smallskip}
		\multirow{4}{*}{TVAE+FMSA} & はるのよの−\underline{ありあけ}のつき−かげもなし \\
		& なほ\underline{ありあけ}の−\underline{ありあけ}のそら \\
		& Spring night, in the dawn sky, waning moon loses the light. \\
		& Silent dawn as ever, the silent dawn sky. \\
		\noalign{\smallskip}\hline\noalign{\smallskip}
		\multirow{4}{*}{RNN-VAD} & うめのはな−ちりくるほどは−なけれども \\
		& はるをのこして−すぎぬべらなり \\
		& Wintersweet, blooming as always. \\
		& Turn around, left Spring behind. \\
		\noalign{\smallskip}\hline\noalign{\smallskip}
		\multirow{4}{*}{WakaVT+FMSA} & みよしのの−やまほととぎす−ながきよの \\
		& やまのみやこの−はるをまつかな \\
		& Wait, Cuckoos on Yoshinoyama, following the lingering night, \\
		& is awaked capital in full spring. \\
		\noalign{\smallskip}\hline
	\end{tabularx}
\end{table}

In order to better illustrate the impact of latent variables on the generated poems, we input the keyword はる(spring) into TLM+FMSA, TVAE+FMSA, RNN-VAD, and WakaVT+FMSA, and then compare the generated results. Table ~\ref{tab:4} shows the generation of each model.

\begin{table}[htbp]
	\caption{Samples generated by WakaVT+FMSA and the corresponding comments on them. The keyword used to generate each poem is underlined.}
	\label{tab:5}
	\begin{tabularx}{\linewidth}{X}
		\hline\noalign{\smallskip}
		\makecell[c]{あけてゆく−みねのこのはの−こずゑより−はるかにつづく−さをしかの\underline{こゑ}} \\
		\makecell[c]{The morning light climbs the treetops of the top mountain,} \\
		\makecell[c]{and the \underline{cry} of stags spreads far and wide.} \\
		\begin{enumerate}[label={reviewer \arabic*}, leftmargin=6em]
			\item The scenery is \emph{fresh with a wild imagination}, and the connection of several images is ingenious.
			\item This poem is \emph{thought-provoking} as it gradually transitions from visual to auditory. The far-reaching cry of the stag can further arouse the longing for the loved one. It is indeed a classical Waka poem.
			\item This Waka is \emph{beautiful as well as touching} under the background of autumn. The author expresses parting sadness through the auditory perspective of the stag's courtship sound.
		\end{enumerate} \\
		\noalign{\smallskip}\hline\noalign{\smallskip}
		\makecell[c]{こひしさは−ひとのこころに−さよふけて−わが\underline{なみだ}こそ−おもひしらるれ} \\
		\makecell[c]{Night approaches, my love is longing in his heart faraway,} \\
		\makecell[c]{my \underline{tears} come and see by all.} \\
		\begin{enumerate}[label={reviewer \arabic*}, leftmargin=6em]
			\item The expression is \emph{proficient and sophisticated with strong feeling}. There is a jump in content between the first two phrases and the last three phrases.
			\item This poem is \emph{grammatically coherent and accurate}, which reminds us of the primal「恋しさに思ひみだれてねぬる夜の深き夢ぢをうつうともがな」, a sleepless night troubled by love.
			\item This poem is \emph{simple in language and sincere in emotion}. The poet depicts the pain of lovesickness and reveals his infinite sadness for incomplete love.
		\end{enumerate} \\
		\noalign{\smallskip}\hline\noalign{\smallskip}
		\makecell[c]{ふるさとの−あとをたづねて−なつくさの−しげみにかかる−をのの\underline{かよひぢ}} \\
		\makecell[c]{I trace the road of my hometown,} \\
		\makecell[c]{the lush grass of summer covers the return \underline{path} in the wilderness.} \\
		\begin{enumerate}[label={reviewer \arabic*}, leftmargin=6em]
			\item The poet’s longing for home is reflected. It is quite \emph{interesting} to know how the fourth phrase connects the third and fifth phrases.
			\item Lush summer grass covered the trace of the hometown road. The entire poem is \emph{closely connected and understandable in meaning}. The accurate grasp of the image of summer grass is quite touching.
			\item This poem is composed with plain language, yet it conveys authentic emotion. It is \emph{simple in style, beautiful in artistic conception, and fluent in language}. The poet truly expresses profound nostalgia.
		\end{enumerate} \\
		\noalign{\smallskip}\hline\noalign{\smallskip}
		\makecell[c]{はてもなき−まがきのくさは−おく\underline{つゆ}に−おもひあまりて−やどるつきかな} \\
		\makecell[c]{Endless weeds by the fence are covered with \underline{dew},} \\
		\makecell[c]{yearning was reflected on the dew by moonlight.} \\
		\begin{enumerate}[label={reviewer \arabic*}, leftmargin=6em]
			\item The idea of connecting certain images such as weeds, dew, and moon is quite \emph{intriguing}. The fourth phrase is \emph{affectionate}.
			\item The weeds beside the fence correspond to melancholy, and dew corresponds to the moon. This poem is \emph{quite classical and interesting}, and it feels like \emph{Kokinshu}.
			\item This Waka is \emph{beautiful in artistic conception}. The dew and the moon's mutual reflection presents an ethereal and tranquil scene, revealing the poet’s romantic feelings of nature.
		\end{enumerate} \\
		\noalign{\smallskip}\hline
	\end{tabularx}
\end{table}

The Waka generated by the model without latent variables, TLM+FMSA, has actually borrowed the first two phrases from the Waka created by human beings\footnote{In the training set, at least 5 poems created by humans start with はる の きる − かすみ の ころも.}. However, the fourth phrase is a repetition of the second phrase, thus the poem is all about rosy clouds from the very beginning to the end, lacking the content development. Although a single latent variable is used in TVAE+FMSA, the content is still relatively simple. The word ありあけ(dawn) was used three times in the generated Waka, destroying the content's integrity. In contrast, the Waka generated by the models with a sequence of latent variables, including RNN-VAD and WakaVT+FMSA, successfully avoided meaningless repetition of words and phrases. By associating はる(spring) with うめのはな(the plum blossom, which is to be withered), RNN-VAD depicts a picture of early spring. WakaVT+FMSA associates the word はる(spring) with ほととぎす(cuckoos), やま(mountain), and みやこ(capital). The description of the cuckoos waiting for the arrival of spring in the capital on Yoshino Mountain diversifies the intended meaning of はる(spring). Moreover, it’s creative to combine やま(mountain), representing the beauty of the nature, and みやこ(capital), representing power and prosperity, in a single phrase. It can be seen that a sequence of latent variables can optimize the model to generate Waka with diversified words, rich content and novel ideas. In contrast, models with a single latent variable or without latent variables may cause word repetition, resulting in meaningless content.

More samples generated by WakaVT+FMSA are shown in Table ~\ref{tab:5}. We invited 3 experts to evaluate each poem individually. Their comments confirmed that WakaVT+FMSA could indeed generate ingenious and innovative Waka poems.

\section{Conclusion and future work}
\label{sec:5}
In this paper, we present a novel Waka generation model, WakaVT, which combines advantages of the latent variable model and the self-attention mechanism. Specifically, a sequence of latent variables is incorporated to model word-level variability in Waka data. Moreover, the proposed self-attention based mechanism FMSA is adopted to learn the hierarchical feature of Waka. Experimental results show that a sequence of latent variables significantly improves the novelty and diversity of the generated poems, and FMSA effectively promotes the linguistic quality in terms of fluency, coherence, and meaningfulness. Considering the objective and subjective evaluation results, our model has apparent advantages over the baselines. In future works, we will apply pretraining methods with large-scale classical literary datasets to make the model better learn the semantic meanings of ancient texts, which we believe can further improve the linguistic quality of generated Waka.

\end{CJK}

\bibliographystyle{unsrtnat}
\bibliography{references}  %%% Uncomment this line and comment out the ``thebibliography'' section below to use the external .bib file (using bibtex) .

%%% Uncomment this section and comment out the \bibliography{references} line above to use inline references.
% \begin{thebibliography}{1}

% 	\bibitem{kour2014real}
% 	George Kour and Raid Saabne.
% 	\newblock Real-time segmentation of on-line handwritten arabic script.
% 	\newblock In {\em Frontiers in Handwriting Recognition (ICFHR), 2014 14th
% 			International Conference on}, pages 417--422. IEEE, 2014.

% 	\bibitem{kour2014fast}
% 	George Kour and Raid Saabne.
% 	\newblock Fast classification of handwritten on-line arabic characters.
% 	\newblock In {\em Soft Computing and Pattern Recognition (SoCPaR), 2014 6th
% 			International Conference of}, pages 312--318. IEEE, 2014.

% 	\bibitem{hadash2018estimate}
% 	Guy Hadash, Einat Kermany, Boaz Carmeli, Ofer Lavi, George Kour, and Alon
% 	Jacovi.
% 	\newblock Estimate and replace: A novel approach to integrating deep neural
% 	networks with existing applications.
% 	\newblock {\em arXiv preprint arXiv:1804.09028}, 2018.

% \end{thebibliography}

\end{document}